# Cue Phrase Classification Using Machine Learning

**Diane J. Litman**                                    DIANE@RESEARCH.ATT.COM

*AT&T Labs - Research, 600 Mountain Avenue*
*Murray Hill, NJ 07974 USA*

## Abstract

Cue phrases may be used in a *discourse* sense to explicitly signal discourse structure, but also in a *sentential* sense to convey semantic rather than structural information. Correctly classifying cue phrases as discourse or sentential is critical in natural language processing systems that exploit discourse structure, e.g., for performing tasks such as anaphora resolution and plan recognition. This paper explores the use of machine learning for classifying cue phrases as discourse or sentential. Two machine learning programs (CGRENDEL and C4.5) are used to induce classification models from sets of pre-classified cue phrases and their features in text and speech. Machine learning is shown to be an effective technique for not only automating the generation of classification models, but also for improving upon previous results. When compared to manually derived classification models already in the literature, the learned models often perform with higher accuracy and contain new linguistic insights into the data. In addition, the ability to automatically construct classification models makes it easier to comparatively analyze the utility of alternative feature representations of the data. Finally, the ease of retraining makes the learning approach more scalable and flexible than manual methods.

## 1. Introduction

*Cue phrases* are words and phrases that may sometimes be used to explicitly signal discourse structure in both text and speech. In particular, when used in a *discourse* sense, a cue phrase explicitly conveys structural information. When used in a *sentential* sense, a cue phrase instead conveys semantic rather than structural information. The following examples (taken from a spoken language corpus that will be described in Section 2) illustrate sample discourse and sentential usages of the cue phrases "say" and "further":

- Discourse

    "...we might have the concept of *say* a researcher who has worked for fifteen years on a certain project ..."
    "*Further*, and this is crucial in AI and probably for expert databases as well ..."

- Sentential

    "...let me just *say* that it bears a strong resemblance to much of the work that's done in semantic nets and even frames."
    "...from a place that is even stranger and *further* away ..."

For example, when used in the discourse sense, the cue phrase "say" conveys the structural information that an example is beginning. When used in the sentential sense, "say" does not convey any structural information and instead functions as a verb.

The ability to correctly classify cue phrases as discourse or sentential is critical for natural language processing systems that need to recognize or convey discourse structure, for tasks such as improving anaphora resolution (Grosz & Sidner, 1986; Reichman, 1985). Consider the following example, again taken from the corpus that will be described in Section 2[1]:

> If the system attempts to hold rules, *say* as an expert database for an expert system, *then* we expect it not only to hold the rules but to in fact apply them for us in appropriate situations.

In this example, the cue phrases "say" and "then" are discourse usages, and explicitly signal the boundaries of an intervening subtopic in the discourse structure. Furthermore, the referents of the noun phrases "the system," "an expert database," and "an expert system" are all possible referents for the pronoun "it." With the structural information conveyed by the cue phrases, the system can determine that "the system" is more relevant for interpreting the pronoun "it," as both "an expert database" and "an expert system" occur within the embedded (and now concluded) subtopic. Without the cue phrases, the reasoning required to determine that the referent of the "the system" is the intended referent of "it" would be much more complex.

Correctly classifying cue phrases as discourse or sentential is important for other natural language processing tasks as well. The discourse/sentential distinction can be used to improve the naturalness of synthetic speech in text-to-speech systems (Hirschberg, 1990). Text-to-speech systems generate synthesized speech from unrestricted text. If a cue phrase can be classified as discourse or sentential using features of the input text, it can then be synthesized using different intonational models for the discourse and sentential usages. In addition, by explicitly identifying rhetorical and other relationships, discourse usages of cue phrases can be used to improve the coherence of multisentential texts in natural language generation systems (Zuckerman & Pearl, 1986; Moser & Moore, 1995). Cue phrases can also be used to reduce the complexity of discourse processing in such areas as argument understanding (Cohen, 1984) and plan recognition (Litman & Allen, 1987; Grosz & Sidner, 1986).

While the problem of cue phrase classification has often been noted (Grosz & Sidner, 1986), until recently, models for classifying cue phrases were neither developed nor evaluated based on careful empirical analyses. Even though the literature suggests that some features might be useful for cue phrase classification, there are no quantitative analyses of any actual classification algorithms that use such features (nor any suggestions as to how different types of features might be combined). Most systems that recognize or generate cue phrases simply assume that discourse uses are utterance or clause initial (Reichman, 1985; Zuckerman & Pearl, 1986). While there are empirical studies showing that the intonational prominence of certain word classes varies with respect to discourse function (Halliday & Hassan, 1976; Altenberg, 1987), these studies do not investigate cue phrases per se.

To address these limitations, Hirschberg and Litman (1993) conducted several empirical studies specifically addressing cue phrase classification in text and speech. Hirschberg and Litman pre-classified a set of naturally occurring cue phrases, described each cue phrase in terms of prosodic and textual features (the features were posited in the literature or easy to automatically code), then manually examined the data to construct classification models that best predicted the classifications from the feature values.

---

1. This example is also described in more detail by Hirschberg and Litman (1993).

This paper examines the utility of machine learning for automating the construction of models for classifying cue phrases from such empirical data. A set of experiments are described that use two machine learning programs, CGRENDEL (Cohen, 1992, 1993) and C4.5 (Quinlan, 1993), to induce classification models from sets of pre-classified cue phrases and their features. The features, classes and training examples used in the studies of Hirschberg and Litman (1993), as well as additional features, classes and training examples, are given as input to the machine learning programs. The results are evaluated both quantitatively and qualitatively, by comparing both the error rates and the content of the manually derived and learned classification models. The experimental results show that machine learning is indeed an effective technique, not only for automating the generation of classification models, but also for improving upon previous results. The accuracy of the learned classification models is often higher than the accuracy of the manually derived models, and the learned models often contain new linguistic implications. The learning paradigm also makes it easier to compare the utility of different knowledge sources, and to update the model given new features, classes, or training data.

The next section summarizes previous work on cue phrase classification. Section 3 then describes the machine learning approach to cue phrase classification that is taken in this paper. In particular, the section describes four sets of experiments that use machine learning to automatically induce cue phrase classification models. The types of inputs and outputs of the machine learning programs are presented, as are the methodologies that are used to evaluate the results. Section 4 presents and discusses the experimental results, and highlights the many benefits of the machine learning approach. Section 5 discusses the practical utility of the results of this paper. Finally, Section 6 discusses the use of machine learning in other studies of discourse, while Section 7 concludes.

## 2. Previous Work on Classifying Cue Phrases

This section summarizes Hirschberg's and Litman's empirical studies of the classification of cue phrases in speech and text (Hirschberg & Litman, 1987, 1993; Litman & Hirschberg, 1990). Hirschberg's and Litman's data (cue phrases taken from corpora of recorded and transcribed speech, classified as *discourse* or *sentential*, and coded using both speech-based and text-based features) will be used to create the input for the machine learning experiments. Hirschberg's and Litman's results (performance figures for manually developed cue phrase classification models) will be used as a benchmark for evaluating the performance of the classification models produced by machine learning.

The first study by Hirschberg and Litman investigated usage of the cue phrase "now" by multiple speakers in a radio call-in show (Hirschberg & Litman, 1987). A classification model based on prosodic features was developed based on manual analysis of a "training" set of 48 examples of "now", then evaluated on a previously unseen test set of 52 examples of "now". In a follow-up study (Hirschberg & Litman, 1993), Hirschberg and Litman tested this classification model on a larger set of cue phrases, namely all single word cue phrases in a technical keynote address by a single speaker. This corpus yielded 953 instances of 34

<u>Prosodic Model:</u>

**if** composition of intermediate phrase = alone **then** *discourse*　　　　(1)
**elseif** composition of intermediate phrase = ¬alone **then**　　　　(2)
　　**if** position in intermediate phrase = first **then**　　　　(3)
　　　　**if** accent = deaccented **then** *discourse*　　　　(4)
　　　　**elseif** accent = L* **then** *discourse*　　　　(5)
　　　　**elseif** accent = H* **then** *sentential*　　　　(6)
　　　　**elseif** accent = complex **then** *sentential*　　　　(7)
　　**elseif** position in intermediate phrase = ¬first **then** *sentential*　　　　(8)

<u>Textual Model:</u>

**if** preceding orthography = true **then** *discourse*　　　　(9)
**elseif** preceding orthography = false **then** *sentential*　　　　(10)

Figure 1: Decision tree representation of the manually derived classification models of Hirschberg and Litman.

different single word cue phrases derived from the literature.[2] Hirschberg and Litman also used the cue phrases in the first 17 minutes of this corpus to develop a complementary cue phrase classification model based on textual features (Litman & Hirschberg, 1990), which they then tested on the full corpus (Hirschberg & Litman, 1993). The first study will be referred to as the "now" study, and the follow-up study as the "multiple cue phrase" study. Note that the term "multiple" means that 34 different single word cue phrases (as opposed to just the cue phrase "now") are considered, not that cue phrases consisting of multiple words (e.g. "by the way") are considered.

The method that Hirschberg and Litman used to develop their prosodic and textual classification models was as follows. They first separately classified each example cue phrase in the data as *discourse*, *sentential* or *ambiguous* while listening to a recording and reading a transcription.[3] Each example was also described as a set of prosodic and textual features.[4] Previous observations in the literature correlating discourse structure with prosodic information, and discourse usages of cue phrases with initial position in a clause, contributed to the choice of features. The set of classified and described examples was then examined in order to manually develop the classification models shown in Figure 1. These models are shown here using decision trees for ease of comparison with the results of C4.5 and will be explained below.

*Prosody* was described using Pierrehumbert's theory of English intonation (Pierrehumbert, 1980). In Pierrehumbert's theory, intonational contours are described as sequences of low (L) and high (H) *tones* in the *fundamental frequency (F0) contour* (the physical correlate of pitch). Intonational contours have as their domain the intonational phrase. A finite-state grammar describes the set of tonal sequences for an intonational phrase. A well-formed *intonational phrase* consists of one or more intermediate phrases followed by a boundary tone. A well-formed *intermediate phrase* has one or more pitch accents followed by a phrase accent. *Boundary tones* and *phrase accents* each consist of a single tone, while *pitch accents* consist of either a single tone or a pair of tones. There are two simple pitch accents (H* and L*) and four *complex* accents (L*+H, L+H*, H*+L, and H+L*). The * indicates which tone is aligned with the stressed syllable of the associated lexical item. Note that not every stressed syllable is accented. Lexical items that bear pitch accents are called *accented*, while those that do not are called *deaccented*.

---

2. Figure 2 contains a list of the 34 cue phrases. Hirschberg and Litman (1993) provide full details regarding the distribution of these cue phrases. The most frequent cue phrase is "and", which occurs 320 times. The next most frequent cue phrase is "now", which occurs 69 times. "But," "like," "or" and "so" also each occur more than fifty times. The four least frequent cue phrases – "essentially," "otherwise," "since" and "therefore" – each occur 2 times.

3. The class *ambiguous* was not introduced until the multiple cue phrase study (Hirschberg & Litman, 1993; Litman & Hirschberg, 1990).

4. Although a limited set of textual features were noted in the "now" data, the analysis of the "now" data did not yield a textual classification model.

Prosody was manually transcribed by Hirschberg by examining the fundamental frequency (F0) contour, and by listening to the recording. This transcription process was performed separately from the process of discourse/sentential classification. To produce the F0 contour, the recording of the corpus was digitized and pitch-tracked using speech analysis software. This resulted in a display of the F0 where the x-axis represented time and the y-axis represented frequency in Hz. Various phrase final characteristics (e.g., phrase accents, boundary tones, as well as pauses and syllable lengthening) helped to identify intermediate and intonational phrases, while peaks or valleys in the display of the F0 contour helped to identify pitch accents. Similar manual transcriptions of prosodic phrasing and accent have been shown to be reliable across coders (Pitrelli, Beckman, & Hirschberg, 1994).

Once prosody was coded, Hirschberg and Litman represented every cue phrase in terms of the following prosodic features.[5] *Accent* corresponded to the pitch accent (if any) that was associated with the cue phrase. For both the intonational and intermediate phrases containing each cue phrase, the feature *composition of phrase* represented whether or not the cue phrase was *alone* in the phrase (the phrase contained only the cue phrase, or only the cue phrase and other cue phrases). *Position in phrase* represented whether the cue phrase was *first* (the first lexical item in the prosodic phrase unit – possibly preceded by other cue phrases) or not.

The *textual features* used in the multiple cue phrase study (Hirschberg & Litman, 1993; Litman & Hirschberg, 1990) were extracted automatically from the transcript. The *part of speech* of each cue phrase was obtained by running a program for tagging words with one of approximately 80 parts of speech (Church, 1988) on the transcript.[6] Several characteristics of the cue phrase's immediate context were also noted, in particular, whether it was immediately preceded or succeeded by *orthography* (punctuation or a paragraph boundary), and whether it was immediately preceded or succeeded by a lexical item corresponding to *another cue phrase.*

With this background, the classification models shown in Figure 1 can now be explained. The prosodic model uniquely classifies any cue phrase using the features *composition of intermediate phrase, position in intermediate phrase,* and *accent.* When a cue phrase is uttered as a single intermediate phrase – possibly with other cue phrases (i.e., line (1) in Figure 1), or in a larger intermediate phrase with an initial position (possibly preceded by other cue phrases) and a L* accent or deaccented, it is classified as *discourse*. When part of a larger intermediate phrase and either in initial position with a H* or complex accent, or in a non-initial position, it is *sentential*. The textual model classifies cue phrases using only the single feature *preceding orthography*.[7] When a cue phrase is preceded by any type of orthography, it is classified as *discourse*; otherwise, the cue phrase is classified as *sentential*.

---

5. Only the features used in Figure 1 are discussed here.
6. Another syntactic feature - dominating constituent - was obtained by running the parser Fidditch (Hindle, 1989) on the transcript. However, since this feature did not appear in any models manually derived from the training data (Litman & Hirschberg, 1990), the feature was not pursued.

| Model | Classifiable Cue Phrases (N=878) | Classifiable Non-Conjuncts (N=495) |
|---|---|---|
| Prosodic | 24.6 ± 3.0 | 14.7 ± 3.2 |
| Textual | 19.9 ± 2.8 | 16.1 ± 3.4 |
| Default Class | 38.8 ± 3.2 | 40.8 ± 4.4 |

Table 1: 95% confidence intervals for the error rates (%) of the manually derived classification models of Hirschberg and Litman, testing data (multiple cue phrase corpus).

When the prosodic model was used to classify each cue phrase in its training data, i.e., the 100 examples of "now" from which the model was developed, the error rate was 2.0%.[8] The error rate of the textual model on the training examples from the multiple cue phrase corpus was 10.6% (Litman & Hirschberg, 1990).

The prosodic and textual models were evaluated by quantifying their performance in correctly classifying example cue phrases in two test sets of data, as shown in the rows labeled "Prosodic" and "Textual" in Table 1. Each test set is a subset of the 953 examples from the multiple cue phrase corpus. The first test set (878 examples) consists of only the *classifiable cue phrases*, i.e., the cue phrases that both Hirschberg and Litman classified as *discourse* or that both classified as *sentential*. Note that those cue phrases that Hirschberg and Litman classified as *ambiguous* or that they were unable to agree upon are not included in the classifiable subset. (These cue phrases will be considered in the learning experiments described in Section 4.4, however.) The second test set, the *classifiable non-conjuncts* (495 examples), was created from the classifiable cue phrases by removing all instances of "and", "or" and "but". This subset was considered particularly reliable since 97.2% of non-conjuncts were classifiable compared to 92.1% of all example cue phrases. The error rate of the prosodic model was 24.6% for the classifiable cue phrases and 14.7% for the classifiable non-conjuncts (Hirschberg & Litman, 1993). The error rate of the textual model was 19.9% for the classifiable cue phrases and 16.1% for the classifiable non-conjuncts (Hirschberg & Litman, 1993). The last row of the table shows error rates for a simple "Default Class" baseline model that always predicts the most frequent class in the corpus (*sentential*). These rates are 38.8% for the classifiable cue phrases and 40.8% for the classifiable non-conjuncts.

---

7. A classification model based on part-of-speech was also developed (Litman & Hirschberg, 1990; Hirschberg & Litman, 1993); however, it did not perform as well as the model based on orthography (the error rate of the part-of-speech model was 36.1% in the larger test set, as opposed to 19.9% for the orthographic model). Furthermore, a model that combined orthography and part-of-speech performed comparably to the simpler orthographic model (Hirschberg & Litman, 1993). Hirschberg and Litman also had preliminary observations suggesting that adjacency of cue phrases might prove useful.

8. Following Hirschberg and Litman (1993), the original 48- and 52-example sets (Hirschberg & Litman, 1987) are combined.

Although not computed by Hirschberg and Litman, Table 1 also associates margins of errors with each error percentage, which are used to compute confidence intervals (Freedman, Pisani, & Purves, 1978). (The margin of error is $\pm$ 2 standard errors for a 95% confidence interval using a normal table.) The lower bound of a confidence interval is computed by subtracting the margin of error from the error rate, while the upper bound is computed by adding the margin of error. Thus, the 95% confidence interval for the prosodic model on the classifiable cue phrase test set is (21.6%, 27.6%). Analysis of the confidence intervals indicates that the improvement of both the prosodic and textual models over the default model is significant. For example, the upper bounds of the error rates of the prosodic and textual models on the classifiable cue phrase test set - 27.6% and 22.7% - are both lower than the lower bound of the default class error rate - 35.6%. This methodology of using statistical inference to determine whether differences in error rates are significant is discussed more fully in Section 3.3.

## 3. Experiments using Machine Learning

This section describes experiments that use the machine learning programs C4.5 (Quinlan, 1993) and CGRENDEL (Cohen, 1992, 1993) to automatically induce cue phrase classification models. CGRENDEL and C4.5 are similar to each other and to other learning methods such as neural networks and CART (Brieman, Friedman, Olshen, & Stone, 1984) in that all induce classification models from preclassified examples. Each program takes the following inputs: names of the classes to be learned, names and possible values of a fixed set of features, and the training data (i.e., a set of examples for which the class and feature values are specified). The output of each program is a classification model, expressed in C4.5 as a decision tree and in CGRENDEL as an ordered set of if-then rules. Both CGRENDEL and C4.5 learn the classification models using greedy search guided by an "information gain" metric.

The first group of machine learning experiments replicate the training and testing conditions used by Hirschberg and Litman (1993) (reviewed in the previous section), to support a direct comparison of the manual and machine learning approaches. The second group of experiments evaluate the utility of training from larger amounts of data than was feasible for the manual analysis of Hirschberg and Litman. The third set of experiments allow the machine learning algorithms to distinguish among the 34 cue phrases, to evaluate the utility of developing classification models specialized for particular cue phrases. The fourth set of experiments consider all the examples in the multiple cue phrase corpus, not just the classifiable cue phrases. This set of experiments attempt to predict a third classification *unknown*, as well as the classifications *discourse* and *sentential*. Finally, within each of these four sets of experiments, each individual experiment learns a classification model using a different feature representation of the training data. Some experiments consider features in isolation, to comparatively evaluate the utility of each individual feature for classification. Other experiments consider linguistically motivated sets of features, to gain insight into feature interactions.

### 3.1 The Machine Learning Inputs

This section describes the inputs to both of the machine learning programs, namely, the names of the classifications to be learned, the names and possible values of a fixed set of features, and training data specifying the class and feature values for each example in the training set.

| Classification | Total | Classifiable Cue Phrases | | Unknown | | | | | | |
|---|---|---|---|---|---|---|---|---|---|---|
| | | Discourse | Sentential | | | | | | | |
| Judge1/Judge2 | | D/D | S/S | ?/? | D/S | S/D | D/? | S/? | ?/D | ?/S |
| All Cue Phrases | 953 | 341 | 537 | 59 | 5 | 0 | 0 | 0 | 5 | 6 |
| Non-Conjuncts | 509 | 202 | 293 | 11 | 1 | 0 | 0 | 0 | 0 | 2 |

Table 2: Determining the classification of cue phrases.

### 3.1.1 CLASSIFICATIONS

The first input to each learning program specifies the names of a fixed set of *classifications*. Hirschberg and Litman's 3-way classification of cue phrases by 2 judges (Hirschberg & Litman, 1993) is transformed into the classifications used by the machine learning programs as shown in Table 2. Recall from Section 2 that each judge classified each cue phrase as *discourse*, *sentential*, or *ambiguous*; these classifications are shown as D, S, and ? in Table 2. As discussed in Section 2, the *classifiable cue phrases* are those cue phrases that the judges both classified as either discourse or as sentential usages. Thus, in the machine learning experiments, a cue phrase is assigned the classification *discourse* if both judges classified it as discourse (D/D, as shown in column 3 of Table 2). Similarly, a cue phrase is assigned the classification *sentential* if both judges classified it as sentential (S/S, as shown in column 4). 878 (92.1%) of the 953 examples in the full corpus were classifiable, while 495 (97.2%) of the 509 non-conjuncts were classifiable.

For some of the machine learning experiments, a third cue phrase classification will also be considered. In particular, a cue phrase is assigned the classification *unknown* if both Hirschberg and Litman classified it as *ambiguous* (?/?, as shown in column 5), or if they were unable to agree upon its classification (D/S, S/D, D/?, S/?, ?/D, ?/S, as shown in columns 6-11). In the full corpus, 59 cue phrases (6.2%) were judged ambiguous by both judges (?/?). There were only 5 cases (.5%) of true disagreement (D/S). 11 cue phrases (1.2%) were judged ambiguous by the first judge but classified by the second judge (?/D and ?/S). When the conjunctions "and," "or" and "but" were removed from the corpus, only 11 examples (2.2%) were judged ambiguous by both judges: 3 instances of "actually," 2 instances each of "because" and "essentially," and 1 instance of "generally," "indeed," "like" and "now." There was only 1 case (.2%) of true disagreement (an instance of "like"). 2 cue phrases (.4%) - an instance each of "like" and "otherwise" - were judged ambiguous by the first judge.

### 3.1.2 FEATURES

A second component of the input to each learning program specifies the names and potential values of a fixed set of *features*. The set of primitive features considered in the learning experiments are shown in Figure 2. Feature values can either be a numeric value or one of a fixed set of user-defined symbolic values. The feature representation shown here follows the representation of Hirschberg and Litman except as noted. *Length of intonational phrase* (P-

- **Prosodic Features**

  - length of intonational phrase (P-L): integer.
  - position in intonational phrase (P-P): integer.
  - length of intermediate phrase (I-L): integer.
  - position in intermediate phrase (I-P): integer.
  - composition of intermediate phrase (I-C): only, only cue phrases, other.
  - accent (A): H*, L*, L*+H, L+H*, H*+L, H+L*, deaccented, ambiguous.
  - accent* (A*): H*, L*, complex, deaccented, ambiguous.

- **Textual Features**

  - preceding cue phrase (C-P): true, false, NA.
  - succeeding cue phrase (C-S): true, false, NA.
  - preceding orthography (O-P): comma, dash, period, paragraph, false, NA.
  - preceding orthography* (O-P*): true, false, NA.
  - succeeding orthography (O-S): comma, dash, period, false, NA.
  - succeeding orthography* (O-S*): true, false, NA.
  - part-of-speech (POS): article, coordinating conjunction, cardinal numeral, subordinating conjunction, preposition, adjective, singular or mass noun, singular proper noun, intensifier, adverb, verb base form, NA.

- **Lexical Feature**

  - token (T): actually, also, although, and, basically, because, but, essentially, except, finally, first, further, generally, however, indeed, like, look, next, no, now, ok, or, otherwise, right, say, second, see, similarly, since, so, then, therefore, well, yes.

Figure 2: Representation of features, for use by C4.5 and CGRENDEL.

L) and *length of intermediate phrase* (I-L) represent the number of words in the intonational and intermediate phrases containing the cue phrase, respectively. This feature was not coded in the "now" data, but was coded (although not used) in the later multiple cue phrase data. *Position in intonational phrase* (P-P) and *position in intermediate phrase* (I-P) use numeric values rather than the earlier symbolic values (e.g., *first* in Figure 1). *Composition of intermediate phrase* (I-C) replaces the value *alone* (meaning that the phrase contained only the example cue phrase, or only the example plus other cue phrases) from Figure 1 with the more primitive values *only* and *only cue phrases* (whose disjunction is equivalent to *alone*); I-C also uses the value *other* rather than ¬*alone* (as was used in Figure 1). *Accent* (A) uses the value *ambiguous* to represent all cases where the prosodic analysis yields a disjunction (e.g., "H*+L or H*"). *Accent** (A*) re-represents some of the symbolic values of the feature *accent* (A) using a more abstract level of description. In particular, L*+H, L+H*, H*+L, and H+L* are represented as separate values in A but as a single value – the superclass *complex* – in A*. While useful abstractions can often result from the learning process, A* is explicitly represented in advance as it is a prosodic feature representation that has the potential to be automated (see Section 5).

In all the textual features, the value *NA* (not applicable) reflects the fact that 39 recorded examples were not included in the transcription, which was done independently of the studies performed by Hirschberg and Litman (1993). In the coding used by Hirschberg and Litman, *preceding cue phrase* (C-P) and *succeeding cue phrase* (C-S) represented the actual cue phrase (e.g., "and") when there was a preceding or succeeding cue phrase; here the value *true* encodes all such cases. As with the prosodic feature set A*, *preceding orthography*\* (O-P*) and *succeeding orthography*\* (O-S*) re-represent some of the symbolic values of *preceding orthography* (O-P) and *succeeding orthography* (O-S), respectively, using a more abstract level of description (e.g., *comma*, *dash*, and *period* are represented as separate values in O-S but as the single value *true* in O-S*). This is done because the reliability of coding detailed transcriptions of orthography is not known. *Part-of-speech* (POS) represents the part of speech assigned to each cue phrase by Church's program for tagging part of speech in unrestricted text (Church, 1988); while the program can assign approximately 80 different values, only the subset of values that were actually assigned to the cue phrases in the transcripts of the corpora are shown in the figure. Finally, the lexical feature *token* (T) is new to this study, and represents the actual cue phrase being described.

### 3.1.3 TRAINING DATA

The final input to each learning program is *training data*, i.e., a set of examples for which the class and feature values are specified. Consider the following utterance, taken from the multiple cue phrase corpus (Hirschberg & Litman, 1993):

**Example 1** [(*Now*) (*now* that we have all been welcomed here)] it's time to get on with the business of the conference.

This utterance contains two cue phrases, corresponding to the two instances of "now". The brackets and parentheses illustrate the intonational and intermediate phrases, respectively, that contain the example cue phrases. Note that a single intonational phrase contains both examples, but that each example is uttered in a different intermediate phrase. If we were only interested in the feature *length of intonational phrase* (P-L), the two examples would be represented in the training data as follows:

| P-L | Class |
|-----|-------|
| 9 | discourse |
| 9 | sentential |

The first column indicates the value assigned to the feature P-L, while the second column indicates how the example was classified. Thus, the length of the intonational phrase containing the first instance of "now" is 9 words, and the example cue phrase is classified as a discourse usage. If we were only interested in the feature *composition of intermediate phrase* (I-C), the two examples would instead be represented in the training data as follows:

| I-C | Class |
|-----|-------|
| only | discourse |
| other | sentential |

That is, the intermediate phrase containing the first instance of "now" contains only the cue phrase "now", while the intermediate phrase containing the second instance of "now" contains "now" as well as 7 other lexical items that are not cue phrases. Note that while the value of P-L is the same for both examples, the value of I-C is different.

## 3.2 The Machine Learning Outputs

The output of both machine learning programs are *classification models.* In C4.5 the model is expressed as a *decision tree,* which consists of either a leaf node (a class assignment), or a decision node (a test on a feature, with one branch and subtree for each possible outcome of the test). The following example illustrates the non-graphical representation for a decision node testing a feature with n possible values:

**if** test$_1$ **then** ...

...

**elseif** test$_n$ **then** ...

Tests are of the form "feature operator value"[9]. "Feature" is the name of a feature (e.g. *accent*), while "value" is a valid value for that feature (e.g., *deaccented*). For features with symbolic values (e.g., *accent*), there is one branch for each symbolic value, and the operator "=" is used. For features with numeric values (e.g., *length of intonational phrase*), there are two branches, each comparing the numeric value with a threshold value; the operators "<" and ">" are used. Given a decision tree, a cue phrase is classified by starting at the root of the tree and following the appropriate branches until a leaf is reached. Section 4 shows example decision trees produced by C4.5.

In CGRENDEL the classification model is expressed as an ordered set of *if-then rules* of the following form:

**if** test$_1$ $\wedge$ ...$\wedge$ test$_k$ **then** *class*

The "if" part of a rule is a conjunction of tests on the values of (varying) features, where tests are again of the form "feature operator value." As in C4.5, "feature" is the name of a feature, and "value" is a valid value for that feature. Unlike C4.5, the operators = or $\neq$ are used for features with symbolic values, while $\leq$ or $\geq$ are used for features with numeric values. The "then" part of a rule specifies a class assignment (e.g, *discourse*). Given a set of if-then rules, a cue phrase is classified using the rule whose "if" part is satisfied. If there or two or more such rules and the rules disagree on the class of an example, CGRENDEL applies one of two conflict resolution strategies (chosen by the user): choose the first rule, or choose the rule that is most accurate on the data. The experiments reported here use the second strategy. If there are no such rules, CGRENDEL assigns a default class. Section 4 shows example rules produced by CGRENDEL.

Both C4.5 and CGRENDEL learn their classification models using greedy search guided by an "information gain" metric. C4.5 uses a divide and conquer process: training examples are recursively divided into subsets (using the tests discussed above), until all of the subsets belong to a single class. The test chosen to divide the examples is that which maximizes a metric called a gain ratio (a local measure of progress, which does not consider any subsequent tests); this metric is based on information theory and is discussed in detail by Quinlan (1993). Once a test is selected, there is no backtracking. Ideally, the set of chosen tests should result in a small final decision tree. CGRENDEL generates its set of if-then rules using a method called *separate and conquer* (to highlight the similarity with divide and conquer):

---

9. An additional type of test may be invoked by a C4.5 option.

> Many rule learning systems generate hypotheses using a greedy strategy in which rules are added to the rule set one by one in an effort to form a small cover of the positive examples; each rule, in turn is created by adding one condition after another to the antecedent until the rule is consistent with the negative data. (Cohen, 1993)

Although CGRENDEL is claimed to have two advantages over C4.5, these advantages do not come into play for the experiments reported here. First, if-then rules appear to be easier for people to understand than decision trees (Quinlan, 1993). However, for the cue phrase classification task, the decision trees produced by C4.5 are quite compact and thus easily understood. Furthermore, a rule representation can be derived from C4.5 decision trees, using the program C4.5rules. Second, CGRENDEL allows users to exploit prior knowledge of a learning problem, by constraining the syntax of the rules that can be learned. However, no prior knowledge is exploited in the cue phrase experiments. The main reason for using both C4.5 and CGRENDEL is to increase the reliability of any comparisons between the machine learning and manual results. In particular, if comparable results are obtained using both C4.5 and CGRENDEL, then any performance differences between the learned and manually derived classification models are less likely to be due to the specifics of a particular learning program, and more likely to reflect the learned/manual distinction.

## 3.3 Evaluation

The output of each machine learning experiment is a classification model that has been learned from the training data. These learned models are qualitatively evaluated by examining their linguistic content, and by comparing them with the manually derived models of Figure 1. The learned models are also quantitatively evaluated by examining their error rates on testing data and by comparing these error rates to each other and to the error rates shown in Table 1. The *error rate* of a classification model is computed by using the model to predict the classifications for a set of examples where the classifications are already known, then comparing the predicted and known classifications. In the cue phrase domain, the error rate is computed by summing the number of discourse examples misclassified as sentential with the number of sentential examples misclassified as discourse, then dividing by the total number of examples.

The error rates of the learned classification models are estimated using two methodologies. *Train-and-test error rate estimation* (Weiss & Kulikowski, 1991) "holds out" a test set of examples, which are not seen until after training is completed. That is, the model is developed by examining only the training examples; the error of the model is then estimated by using the model to classify the test examples. This was the evaluation method used by Hirschberg and Litman. The resampling method of *cross-validation* (Weiss & Kulikowski, 1991) estimates error rate using multiple train-and-test experiments. For example, in 10-fold cross-validation, instead of dividing examples into training and test sets once, 10 runs of the learning program are performed. The total set of examples is randomly divided into 10 disjoint test sets; each run thus uses the 90% of the examples not in the test set for training and the remaining 10% for testing. Note that for each iteration of the cross-validation, the learning process begins from scratch; thus a new classification model is learned from each training sample. An estimated error rate is obtained by averaging the error rate on the test-ing portion of the data from each of the 10 runs. While this method does not make sense for humans, computers can truly ignore previous iterations. For sample sizes in the hundreds (the classifiable subset of the multiple cue phrase sample and the classifiable non-conjunct subset provide 878 and 495 examples, respectively) 10-fold cross-validation often provides a better performance estimate than the hold-out method (Weiss & Kulikowski, 1991). The major advantage is that in cross-validation all examples are eventually used for testing, and almost all examples are used in any given training run.

The best performing learned models are identified by comparing their error rates to the error rates of the other learned models and to the manually derived error rates. To determine whether the fact that an error rate E1 is lower than another error rate E2 is also significant, statistical inference is used. In particular, confidence intervals for the two error rates are computed, at a 95% confidence level. When an error rate is estimated using only a single error rate on a test set (i.e., the train-and-test methodology), the confidence interval is computed using a normal approximation to the binomial distribution (Freedman et al., 1978). When the error rate is estimated using the average from multiple error rates (i.e., the cross-validation methodology), the confidence interval is computed using a $t$-Table (Freedman et al., 1978). If the upper bound of the 95% confidence interval for E1 is lower than the lower bound of the 95% confidence interval for the error rate E2, then the difference between E1 and E2 is assumed to be significant.[10]

## 3.4 The Experimental Conditions

This section describes the conditions used in each set of machine learning experiments. The experiments differ in their use of training and testing corpora, methods for estimating error rates, and in the features and classifications used. The actual results of the experiments are presented in Section 4.

### 3.4.1 Four Sets of Experiments

The learning experiments can be conceptually divided into four sets. Each experiment in the first set estimates error rate using the train-and-test method, where the training and testing samples are those used by Hirschberg and Litman (1993) (the "now" data and the two subsets of the multiple cue phrase corpus, respectively). This allows a direct comparison of the manual and machine learning approaches. However, only the prosodic experiments conducted by Hirschberg and Litman (1993) are replicated. The textual training and testing conditions are not replicated as the original training corpus (the first 17 minutes of the multiple cue phrase corpus) (Litman & Hirschberg, 1990) is a subset of, rather than disjoint from, the test corpus (the full 75 minutes of the multiple cue phrase corpus) (Hirschberg & Litman, 1993).

In contrast, each experiment in the second set uses cross-validation to estimate error rate. Furthermore, both training and testing samples are taken from the multiple cue phrase corpus. Each experiment uses 90% of the examples from the multiple cue phrase data for training, and the remaining 10% for testing. Thus each experiment in the second set trains from much larger amounts of data (790 classifiable examples, or 445 classifiable non-conjuncts) than each experiment in the first set (100 "nows"). The reliability of the testing is not compromised due to the use of cross-validation (Weiss & Kulikowski, 1991).

---

10. Thanks to William Cohen for suggesting this methodology.

| | P-L | P-P | I-L | I-P | I-C | A | A* | C-P | C-S | O-P | O-P* | O-S | O-S* | POS |
|---|---|---|---|---|---|---|---|---|---|---|---|---|---|---|
| prosody | X | X | X | X | X | X | X | | | | | | | |
| hl93features | | | | X | X | X | X | | | | | | | |
| phrasing | X | X | X | X | | | | | | | | | | |
| length | X | | X | | | | | | | | | | | |
| position | | X | | X | | | | | | | | | | |
| intonational | X | X | | | | | | | | | | | | |
| intermediate | | | X | X | X | | | | | | | | | |
| text | | | | | | | | X | X | X | X | X | X | X |
| adjacency | | | | | | | | X | X | | | | | |
| orthography | | | | | | | | | | X | X | X | X | |
| preceding | | | | | | | | X | | X | X | | | |
| succeeding | | | | | | | | | X | | | X | X | |
| speech-text | X | X | X | X | X | X | X | X | X | X | X | X | X | X |

Table 3: Multiple feature sets and their components.

Each experiment in the third set replicates an experiment in the second set, with the exception that the learning program is now allowed to distinguish between cue phrases. This is done by adding a feature representing the cue phrase (the feature *token* from Figure 2) to each experiment from the second set. Since the potential use of such a lexical feature was noted but not used by Hirschberg and Litman (1993), these experiments provide qualitatively new linguistic insights into the data. For example, the same features may now be used differently to predict the classifications of different cue phrases or sets of cue phrases.

Finally, each experiment in the fourth set replicates an experiment in the first, second, and third set, with the exception that all 953 examples in the multiple cue phrase corpus are now considered. This is because in practice, any learned cue phrase classification model will likely be used to classify all cue phrases, even those that are difficult for human judges to classify. The experiments in the fourth set allow the learning programs to attempt to learn the class *unknown*, in addition to the classes *discourse* and *sentential*.

### 3.4.2 Feature Representations within Experiment Sets

Within each of these four sets of experiments, each individual experiment represents the data using a different subset of the available features. First, the data is represented in each of 14 *single feature sets*, corresponding to each prosodic and textual feature shown in Figure 2. These experiments comparatively evaluate the utility of each individual feature for classification. The representations of Example 1 shown above illustrate how data is represented using the single feature set P-L, and using the single feature set I-C.

Second, the data is represented in each of the 13 *multiple feature sets* shown in Table 3. Each of these sets contains a linguistically motivated subset of at least 2 of the 14 features. The first 7 sets use only prosodic features. *Prosody* considers all the prosodic features that were coded for each example cue phrase. *Hl93features* considers only the coded features that were also used in the model shown in Figure 1. *Phrasing* considers all features of both the intonational and intermediate phrases containing the example cue phrase (i.e., length

**Example 1**  [(*Now*) (*now* that we have all been welcomed here)] it's time to get on with the business of the conference.

| P-L | P-P | I-L | I-P | I-C | A | A* | C-P | C-S | O-P | O-P* | O-S | O-S* | POS | Class |
|-----|-----|-----|-----|-----|-----|-----|-----|-----|-----|------|-----|------|-----|-------|
| 9 | 1 | 1 | 1 | only | H*+L | complex | f | t | par. | t | f | f | adv. | disc. |
| 9 | 2 | 8 | 1 | other | H* | H* | t | f | f | f | f | f | adv. | sent. |

Figure 3: Representation of Example 1 in feature set speech-text.

of phrase, position of example in phrase, and composition of phrase). *Length* and *position* each consider only one of these features, but with respect to both the intonational and intermediate phrase. Conversely, *intonational* and *intermediate* each consider only one type of phrase, but consider all of the features. The next 5 sets use only textual features. *Text* considers all the textual features. *Adjacency* and *orthography* each consider a single textual feature, but consider both the preceding and succeeding immediate context. *Preceding* and *succeeding* consider contextual features relating to both orthography and cue phrases, but limit the context. The last set, *speech-text*, uses all of the prosodic and textual features.

Figure 3 illustrates how the two example cue phrases in Example 1 would be represented using *speech-text*. Consider the feature values for the first example cue phrase. Since this example is the first lexical item in both the intonational and intermediate phrases which contain it, its position in both phrases (P-P and I-P) is 1. Since the intermediate phrase containing the cue phrase contains no other lexical items, its length (I-L) is 1 word and its composition (I-C) is *only* the cue phrase. The values for A and A* indicate that when the intonational phrase is described as a sequence of tones, the complex pitch accent H*+L is associated with the cue phrase. With respect to the textual features, the utterance was transcribed such that it began a new paragraph. Thus the example cue phrase was not preceded by another cue phrase (C-P), but it was preceded by a form of orthography (O-P and O-P*). Since the example cue phrase was immediately followed by another instance of "now" in the transcription, the cue phrase was succeeded by another cue phrase (C-S) but was not succeeded by orthography (O-S and O-S*). Finally, the output of the part of speech tagging program when run on the transcript of the corpus yields the value *adverb* for the cue phrase's part of speech (POS).

The first set of experiments replicate only the prosodic experiments conducted by Hirschberg and Litman (1993); cue phrases are represented using the subset of the feature sets that only consist of prosodic features. In the second set of experiments, examples are represented using all 27 different feature sets (the 14 single feature sets and the 13 multiple feature sets). In the third set of experiments, examples are represented using 27 *tokenized feature sets*, constructed by adding the lexical feature *token* from Figure 2 (the cue phrase being described) to each of the 14 single and 13 multiple feature sets from the second set of experiments. These tokenized feature sets will be referred to using the names of the single and multiple feature sets, concatenated with "+". The following illustrates how the two cue phrases in Example 1 would be represented using P-L+:

| P-L | T | Class |
|-----|-----|-------|
| 9 | now | discourse |
| 9 | now | sentential |

The representation is similar to the P-L representation shown earlier, except for the second column which indicates the value assigned to the feature *token* (T).

## 4. Results

This section examines the results of running the two learning programs – C4.5 and CGREN-DEL – in the four sets of cue phrase classification experiments described above. The learned classification models will be compared with the classification models shown in Figure 1, while the error rates of the learned classification models will be compared with the error rates shown in Table 1 and with the error rates of the other learned models. As will be seen, the results suggest that machine learning is useful for automating the generation of linguistically viable classification classification models, for generating classification models that perform with lower error rates than manually developed hypotheses, and for adding to the body of linguistic knowledge regarding cue phrases.

### 4.1 Experiment Set 1: Replicating Hirschberg and Litman

The first group of experiments replicate the training, testing, and evaluation conditions used by Hirschberg and Litman (1993), in order to investigate how well machine learning performs in comparison to the manual development of cue phrase classification models.

Figure 4 shows the best performing prosodic classification models learned by the two machine learning programs; the top of the figure replicates the manually derived prosodic model from Figure 1 for ease of comparison. When all of the prosodic features are used to represent the 100 training examples of "now" (i.e., each example is represented using feature set *prosody* from Table 3)[11], the classification models that are learned are shown after the manually derived model at the top of Figure 4. Note that using both learning programs, the same decision tree is also learned when the smaller feature sets *phrasing* and *position* are used to represent the "now" data. The bottom portion of the figure shows the classification models that are learned when the same examples are represented using only the single prosodic feature *position in intonational phrase* (P-P); the same model is also learned when the examples are represented using the multiple feature set *intonational*.

Recall that C4.5 represents each learned classification model as a decision tree. Each level of the tree (shown by indentation) specifies a test on a single feature, with a branch for every possible outcome of the test. A branch can either lead to the assignment of a class, or to another test. For example, the C4.5 classification model learned from *prosody* classifies cue phrases using the two features *position in intonational phrase* (P-P) and *position in intermediate phrase* (I-P). Note that not all of the available features in *prosody* (recall Table 3) are used in the decision tree. The tree initially branches based on the value of the feature *position in intonational phrase*.[12] The first branch leads to the class assignment *discourse*. The second branch leads to a test of the feature *position in intermediate phrase*. The first branch of this test leads to the class assignment *discourse*, while the second branch leads to *sentential*. C4.5 produces both unsimplified and pruned decision trees. The goal

---

11. In Experiment Set 1, the feature set *prosody* does not contain the features P-L and I-L. Recall that phrasal length was only coded in the later multiple cue phrase study.

12. For ease of comparison to Figure 1, the original symbolic representation of the feature value is used rather than the integer representation shown in Figure 2.

Manually derived prosodic model (repeated from Figure 1):

| | |
|---|---|
| **if** composition of intermediate phrase = alone **then** *discourse* | (1) |
| **elseif** composition of intermediate phrase = ¬alone **then** | (2) |
|     **if** position in intermediate phrase = first **then** | (3) |
|         **if** accent = deaccented **then** *discourse* | (4) |
|         **elseif** accent = L* **then** *discourse* | (5) |
|         **elseif** accent = H* **then** *sentential* | (6) |
|         **elseif** accent = complex **then** *sentential* | (7) |
|     **elseif** position in intermediate phrase = ¬first **then** *sentential* | (8) |

Decision tree learned from prosody, from phrasing, and from position using C4.5:

**if** position in intonational phrase = first **then** *discourse*
**elseif** position in intonational phrase = ¬first **then**
    **if** position in intermediate phrase = first **then** *discourse*
    **elseif** position in intermediate phrase = ¬first **then** *sentential*

Ruleset learned from prosody, from phrasing, and from position using CGRENDEL:

**if** (position in intonational phrase ≠ first) ∧ (position in intermediate phrase ≠ first) **then** *sentential*
default is on *discourse*

Decision tree learned from P-P and from intonational using C4.5:

**if** position in intonational phrase = first **then** *discourse*
**elseif** position in intonational phrase = ¬first **then** *sentential*

Ruleset learned from P-P and from intonational using CGRENDEL:

**if** position in intonational phrase ≠ first **then** *sentential*
default is on *discourse*

Figure 4: Example C4.5 and CGRENDEL classification models learned from different prosodic feature representations of the "now" data.

| Model | Classifiable Cue Phrases (N=878) | Classifiable Non-Conjuncts (N=495) |
|---|---|---|
| P-P | *18.3 ± 2.6* | 16.6 ± 3.4 |
| prosody | 27.3 ± 3.0 | 17.8 ± 3.4 |
| phrasing | 27.3 ± 3.0 | 17.8 ± 3.4 |
| position | 27.3 ± 3.0 | 17.8 ± 3.4 |
| intonational | *18.3 ± 2.6* | 16.6 ± 3.4 |
| manual prosodic | 24.6 ± 3.0 | 14.7 ± 3.2 |

Table 4: 95%-confidence intervals for the error rates (%) of the best performing CGRENDEL prosodic classification models, testing data. (Training data was the "now" corpus; testing data was the multiple cue phrase corpus.)

of the pruning process is to take a complex decision tree that may also be overfitted to the training data, and to produce a tree that is more comprehensible and whose accuracy is not comprised (Quinlan, 1993). Since almost all trees are improved by pruning (Quinlan, 1993), only simplified decision trees are considered in this paper.

In contrast, CGRENDEL represents each learned classification model as a set of if-then rules. Each rule specifies a conjunction of tests on various features, and results in the assignment of a class. For example, the CGRENDEL ruleset learned from *prosody* classifies cue phrases using the two features *position in intonational phrase* (P-P) and *position in intermediate phrase* (I-P) (the same two features used in the C4.5 decision tree). If the values of both features are not *first*, the if-then rule applies and the cue phrase is classified as *sentential*. If the value of either feature is *first*, the default applies and the cue phrase is classified as *discourse*.

An examination of the learned classification models of Figure 4 shows that they are comparable in content to the portion of the manually derived model that classifies cue phrases solely on phrasal position (line (8)). In particular, all of the classification models say that if the cue phrase is not in an initial phrasal position classify it as *sentential*. On the other hand, the manually derived model also assigns the class *sentential* given an initial phrasal position in conjunction with certain combinations of phrasal composition and accent; the learned classification models instead classify the cue phrase as *discourse* in all other cases. As will be shown, the further discrimination of the manually obtained model does not significantly improve performance when compared to the learned classification models, and in fact in one case significantly degrades performance.

The error rates of the learned classification models on the "now" training data from which they were developed is as follows: 6% for the models learned from *prosody*, *phrasing* and *position*, and 9% for the models learned from P-P and *intonational*. Recall from Section 2 that the error rate of the manually developed prosodic model of Figure 1 on the same training data was 2%.

Table 4 presents 95% confidence intervals for the error rates of the best performing CGRENDEL prosodic classification models. For ease of comparison, the row labeled "manual prosodic" presents the error rates of the manually developed prosodic model of Figure 1 on the same two test sets, which were originally shown in Table 1. The table includes all the CGRENDEL models whose performance matches or exceeds the manual performance.

Comparison of the error rates of the learned and manually developed models suggests that machine learning is an effective technique for automating the development of cue phrase classification models. In particular, within each test set, the 95% confidence interval for the error rate of the classification models learned from the multiple feature sets *prosody*, *phrasing*, and *position* each overlaps with the confidence interval for the error rate of the manual prosodic model. This is also true for the error rates of P-P and *intonational* in the classifiable non-conjunct test set. Thus, machine learning supports the automatic construction of a variety of cue phrase classification models that achieve similar performance as the manually constructed models.

The results from P-P and from *intonational* in the classifiable cue phrase test set are shown in italics, as they suggest that machine learning may also be useful for improving performance. Although the very simple classification model learned from P-P and *intonational* performs worse than the manually derived model on the training data, when tested on the classifiable cue phrases, the learned model (with an upper bound error rate of 20.9%) outperforms the manually developed model (with a lower bound error rate of 21.6%). This suggests that the manually derived model might have been overfitted to the training data, i.e., that the prosodic feature set most useful for classifying "now" did not generalize to other cue phrases. As noted above, the use of simplified learned classification models helps to guard against overfitting in the learning approach. The ease of inducing classification models from many different sets of features using machine learning supports the generation and evaluation of a wide variety of hypotheses (e.g. P-P, which was a high performing but not the optimal performing model on the training data).

Note that the manual prosodic manual performs significantly better in the smaller test set (which does not contain the cue phrases "and", "or", and "but"). In contrast, the performance improvement for P-P and *intonational* in the smaller test set is not significant. This also suggests that the manually derived model does not generalize as well as the learned models.

Finally, for the feature sets shown in Table 4, the decision trees produced by C4.5 perform with the same error rates as the rulesets produced by CGRENDEL, for both test sets. Recall from Figure 4 that the C4.5 decision trees and CGRENDEL rules are in fact semantically equivalent for each feature set. The fact that comparable results are obtained using C4.5 and CGRENDEL adds an extra degree of reliability to the experiments. In particular, the duplication of the results suggests that the ability to match and perhaps even to improve upon manual performance by using machine learning is not due to the specifics of either learning program.

## 4.2 Experiment Set 2: Using Different Training Sets

The second group of experiments evaluate the utility of training from larger amounts of data. This is done by using 10-fold cross-validation to estimate error, where for each run 90% of the examples in a sample are used for training (and over the 10 runs, all of the examples are used for testing). In addition, the experiments in this second set take both the training and testing data from the multiple-cue phrase corpus, in contrast to the previous set of experiments where the training data was taken from the "now" corpus. As will be seen, these changes improve the results, such that more of the learned classification models

| Model | Classifiable Cue Phrases (N=878) | Classifiable Non-Conjuncts (N=495) |
|---|---|---|
| P-L | 33.0 ± 5.9 | (33.2 ± 1.9) |
| P-P | *16.1 ± 3.5* | 18.8 ± 4.2 |
| I-L | 25.5 ± 3.7 | (25.6 ± 2.8) |
| I-P | 25.9 ± 4.9 | 19.4 ± 3.1 |
| I-C | (36.5 ± 5.4) | (35.2 ± 3.4) |
| A | 28.6 ± 3.6 | (30.2 ± 3.1) |
| A* | 28.3 ± 4.3 | (28.4 ± 1.7) |
| prosody | *15.5 ± 2.6* | 17.2 ± 3.1 |
| hl93features | 29.4 ± 3.3 | 18.2 ± 4.2 |
| phrasing | *16.1 ± 3.4* | 19.6 ± 3.9 |
| length | 26.1 ± 3.8 | (27.4 ± 3.4) |
| position | *18.2 ± 2.3* | 19.4 ± 2.8 |
| intonational | *17.0 ± 4.0* | 20.6 ± 3.6 |
| intermediate | 21.9 ± 2.3 | 19.4 ± 5.7 |
| manual prosodic | 24.6 ± 3.0 | 14.7 ± 3.2 |

Table 5: 95%-confidence intervals for the error rates (%) of all CGRENDEL prosodic classi-
fication models, testing data. (Training and testing were done from the multiple
cue phrase corpus using cross-validation.)

perform with lower or comparable error rates when compared to the manually developed
models.

### 4.2.1 PROSODIC MODELS

Table 5 presents the error rates of the classification models learned by CGRENDEL, in
the 28 different prosodic experiments. (For Experiment Sets 2 and 3, the C4.5 error rates
are presented in Appendix A.) Each numeric cell shows the 95% confidence interval for the
error rate, which is equal to the error percentage obtained by cross-validation ± the margin
of error (± 2.26 standard errors, using a *t*-Table). The top portion of the table considers
the models learned from the single prosodic feature sets (Figure 2), the middle portion
considers the models learned from the multiple feature sets (Table 3), while the last row
considers the manually developed prosodic model. The error rates shown in italics indicate
that the performance of the learned classification model exceeds the performance of the
manual model (given the same test set). The error rates shown in parentheses indicate the
opposite case - that the performance of the manual model exceeds the performance of the
learned model. Such cases were omitted in Table 4.

As in Experiment Set 1, comparison of the error rates of the learned and manually
developed models suggests that machine learning is an effective technique for not only
automating the development of cue phrase classification models, but also for improving
performance. When evaluated on the classifiable cue phrase test set, five learned models
have improved performance compared to the manual model; all of the models except I-C
perform at least comparably to the manual model. Note that in Experiment Set 1, only two
learned models outperformed the manual model, and only five learned models performed
at least comparably. The ability to use large training sets thus appears to be an advantage
of the automated approach.

Manually derived prosodic model (repeated from Figure 1):

| | |
|---|---|
| **if** composition of intermediate phrase = alone **then** *discourse* | (1) |
| **elseif** composition of intermediate phrase = ¬alone **then** | (2) |
|     **if** position in intermediate phrase = first **then** | (3) |
|         **if** accent = deaccented **then** *discourse* | (4) |
|         **elseif** accent = L* **then** *discourse* | (5) |
|         **elseif** accent = H* **then** *sentential* | (6) |
|         **elseif** accent = complex **then** *sentential* | (7) |
|     **elseif** position in intermediate phrase = ¬first **then** *sentential* | (8) |

Decision tree learned from P-P using C4.5:

**if** position in intonational phrase ≤ 1 **then** *discourse*
**elseif** position in intonational phrase > 1 **then** *sentential*

Ruleset learned from P-P using CGRENDEL:

**if** position in intonational phrase ≥ 2 **then** *sentential*
default is on *discourse*

Decision tree learned from prosody using C4.5:

**if** position in intonational phrase ≤ 1 **then**
    **if** position in intermediate phrase ≤ 1 **then** *discourse*
    **elseif** position in intermediate phrase > 1 **then** *sentential*
**elseif** position in intonational phrase > 1 **then**
    **if** length of intermediate phrase ≤ 1 **then** *discourse*
    **elseif** length of intermediate phrase > 1 **then** *sentential*

Ruleset learned from prosody using CGRENDEL:

**if** (position in intonational phrase ≥ 2) ∧ (length of intermediate phrase ≥ 2) **then** *sentential*
**if** (7 ≥ position in intonational phrase ≥ 4) ∧ (length of intonational phrase ≥ 10) **then** *sentential*
**if** (length of intermediate phrase ≥ 2) ∧ (length of intonational phrase ≤ 7) ∧ (accent = H*) **then** *sentential*
**if** (length of intermediate phrase ≥ 2) ∧ (length of intonational phrase ≤ 9) ∧ (accent = H*+L) **then** *sentential*
**if** (length of intermediate phrase ≥ 2) ∧ (accent = deaccented) **then** *sentential*
**if** (length of intermediate phrase ≥ 8) ∧ (length of intonational phrase ≤ 9) ∧ (accent = L*) **then** *sentential*
default is on *discourse*

Figure 5: Example C4.5 and CGRENDEL classification models learned from different prosodic feature representations of the classifiable cue phrases in the multiple cue phrase corpus.

When tested on the classifiable non-conjuncts (where the error rate of the manually derived model decreases), machine learning is useful for automating but not for improving performance. This might reflect the fact that the manually derived theories already achieve optimal performance with respect to the examined features in this less noisy subcorpus, and/or that the automatically derived theory for this subcorpus was based on a smaller training set than used in the larger subcorpus.

An examination of some of the best performing learned classification models shows that they are quite comparable in content to relevant portions of the prosodic model of Figure 1, and often contain further linguistic insights. Consider the classification model learned from the single feature *position in intonational phrase* (P-P), shown near the top of Figure 5.

Both of the learned classification models say that if the cue phrase is not in the initial position of the intonational phrase, classify as *sentential*; otherwise classify as *discourse*. Note the correspondence with line (8) in the manually derived prosodic model. Also note that the classification models are comparable[13] to the P-P classification models learned from Experiment Set 1 (shown in Figure 4), despite the differences in training data. The fact that the single prosodic feature *position in intonational phrase* (P-P) can classify cue phrases at least as well as the more complicated manual and multiple feature learned models is again a new result of the learning experiments.

Figure 5 also illustrates the more complex classification models learned using *prosody*, the largest prosodic feature set. The C4.5 model is similar to lines (1) and (8) of the manual model. (The length value 1 is equivalent to the composition value *alone*.) In the ruleset induced from *prosody* by CGRENDEL, the first 2 if-then rules correlate *sentential* status with (among other things) non-initial position[14], and the second 2 rules with H* and H*+L accents; these rules are similar to lines (6)-(8) in Figure 1. However, the last 2 if-then rules in the ruleset also correlate no accent and L* with sentential status when the phrase is of a certain length, while lines (4) and (5) in Figure 1 provide a different interpretation and do not take length into account. Recall that length was coded by Hirschberg and Litman only in their test data. Length was thus never used to generate or revise their prosodic model. The utility of length is a new result of this experiment set.

Although not shown, the models learned from *phrasing*, *position*, and *intonational* also outperform the manual model. As can be seen from Table 3, these models correspond to all of the feature sets that are supersets of P-P but subsets of *prosody*.

### 4.2.2 TEXTUAL MODELS

Table 6 presents the error rates of the classification models learned by CGRENDEL, in the 24 different textual experiments. Unlike the experiments involving the prosodic feature sets, none of the learned textual models perform significantly better than the manually derived model. However, the results suggest that machine learning is still an effective technique for automating the development of cue phrase classification models. In particular, five learned models (O-P, O-P*, *text*, *orthography*, and *preceding*) perform comparably to the manually derived model, in both test sets. Note that these five models are learned from the five textual feature sets that include either the feature O-P or O-P* (recall Figure 2 and Table 3). These models perform significantly better than all of the remaining learned textual models.

Figure 6 shows the best performing learned textual models. Note the similarity to the manually derived model. As with the prosodic results, the best performing single feature models perform comparably to those learned from multiple features. In fact, in CGRENDEL, the rulesets learned from the multiple feature sets *orthography* and *preceding* are identical to the rulesets learned from the single features O-P and O-P*, even though more features were available for use. (The corresponding error rates in Table 6 are not identical due to the

---

13. The different feature values in the two figures reflect the fact that phrasal position was represented in the "now" corpus using symbolic values (as in Figure 1), and in the multiple cue phrase corpus using integers (as in Figure 2).

14. Tests such as "feature $\geq$ x" and "feature $\leq$ y" are merged in the figure for simplicity, e.g., "y $\geq$ feature $\geq$ x."

| Model | Classifiable Cue Phrases (N=878) | Classifiable Non-Conjuncts (N=495) |
|---|---|---|
| C-P | (40.7 ± 6.2) | (40.2 ± 4.5) |
| C-S | (41.3 ± 5.9) | (39.8 ± 4.2) |
| O-P | 20.6 ± 5.7 | 17.6 ± 3.3 |
| O-P* | 18.4 ± 3.7 | 17.2 ± 2.4 |
| O-S | (34.1 ± 6.3) | (30.2 ± 1.8) |
| O-S* | (35.2 ± 5.5) | (32.6 ± 3.0) |
| POS | (37.7 ± 4.1) | (38.2 ± 4.6) |
| text | 18.8 ± 4.2 | 19.0 ± 3.6 |
| adjacency | (39.7 ± 5.7) | (40.2 ± 3.4) |
| orthography | 18.9 ± 3.4 | 18.8 ± 3.0 |
| preceding | 18.8 ± 3.8 | 17.6 ± 3.2 |
| succeeding | (33.9 ± 6.0) | (30.0 ± 2.7) |
| manual textual | 19.9 ± 2.8 | 16.1 ± 3.4 |

Table 6: 95%-confidence intervals for the error rates (%) of all CGRENDEL textual classification models, testing data. (Training and testing were done from the multiple cue phrase corpus using cross-validation.)

---

Manually derived textual model (repeated from Figure 1):

**if** preceding orthography = true **then** *discourse*
**elseif** preceding orthography = false **then** *sentential*

---

Decision tree learned from O-P*, from text, from orthography, and from preceding using C4.5:

**if** preceding orthography* = NA **then** *discourse*
**elseif** preceding orthography* = false **then** *sentential*
**elseif** preceding orthography* = true **then** *discourse*

---

Ruleset learned from O-P, from O-P*, from orthography, and from preceding using CGRENDEL:

**if** preceding orthography* = false **then** *sentential*
default is on *discourse*

---

Ruleset learned from text using CGRENDEL:

**if** preceding orthography* = false **then** *sentential*
**if** part-of-speech = article **then** *sentential*
default is on *discourse*

Figure 6: Example C4.5 and CGRENDEL classification models learned from different textual feature representations of the classifiable cue phrases in the multiple cue phrase corpus.

estimation using cross-validation.) The CGRENDEL model *text* also incorporates the feature *part-of-speech*. In C4.5, the models *text*, *orthography* and *preceding* are all identical to O-P*.

### 4.2.3 PROSODIC/TEXTUAL MODELS

Table 7 presents the error rates of the classification models learned by CGRENDEL when the data is represented using *speech-text*, the complete set of prosodic and textual features (recall

| Model | Classifiable Cue Phrases (N=878) | Classifiable Non-Conjuncts (N=495) |
|---|---|---|
| speech-text | *15.9 ± 3.2* | 14.6 ± 4.6 |
| manual prosodic | 24.6 ± 3.0 | 14.7 ± 3.2 |
| manual textual | 19.9 ± 2.8 | 16.1 ± 3.4 |

Table 7: 95%-confidence intervals for the error rates (%) of the CGRENDEL prosodic/textual classification model, testing data. (Training and testing were done from the multiple cue phrase corpus using cross-validation.)

Table 3). Since Hirschberg and Litman did not develop a similar classification model that combined both types of features, for comparison the last two rows show the error rates of the separate prosodic and textual models. Only when the learned model is compared to the manual prosodic model, using the classifiable cue phrases for testing, does learning result in a significant performance improvement. This is consistent with the results discussed above, where several learned prosodic models performed better than the manually derived prosodic model in this test set. The performance of *speech-text* is not significantly better or worse than the performance of either the best prosodic or textual learned models (Tables 5 and 6, respectively).

Figure 7 shows the C4.5 and CGRENDEL hypotheses learned from *speech-text*. The C4.5 model classifies cue phrases using the prosodic and textual features that performed best in isolation (*position in intonational phrase* and *preceding orthography\**, as discussed above), in conjunction with the additional feature *length of intermediate phrase* (which also appears in the model learned from *prosody* in Figure 5). Like line (9) in the manually derived textual model, the learned model associates the presence of preceding orthography with the class *discourse*. Unlike line (10), however, cue phrases not preceded by orthography may be classified as either *discourse* or *sentential*, based on prosodic feature values (which were not available for use by the textual model). The branch of the learned decision tree corresponding to the last three lines is also similar to lines (1), (2), and (8) of the manually derived prosodic model. (Recall that a length value of 1 is equivalent to a composition value *alone*.)

The CGRENDEL model uses similar features to those used by C4.5 as well as the prosodic feature *accent* (also used in *prosody* in Figure 5), and the textual features *part-of-speech* (also used in *text* in Figure 6) and *preceding cue phrase*. Like C4.5, and unlike line (10) of the manually derived textual model, the CGRENDEL model classifies cue phrases lacking preceding orthography as *sentential* only in conjunction with certain other feature values. Unlike line (9) in the manual model, the learned model also classifies some cue phrases with preceding orthography as *sentential* (if the orthography is a comma, and other feature values are present). Finally, the third and fifth learned rules elaborate line (6) with additional prosodic as well as textual features, while the first and last learned rules elaborate line (8).

## 4.3 Experiment Set 3: Adding the Feature *token*

Each experiment in the third group replicates an experiment from the second group, with the exception that the data representation now also includes the lexical feature *token* from

<u>Manually derived prosodic model (repeated from Figure 1):</u>

**if** composition of intermediate phrase = alone **then** *discourse*         (1)
**elseif** composition of intermediate phrase = ¬alone **then**         (2)
    **if** position in intermediate phrase = first **then**         (3)
        **if** accent = deaccented **then** *discourse*         (4)
        **elseif** accent = L* **then** *discourse*         (5)
        **elseif** accent = H* **then** *sentential*         (6)
        **elseif** accent = complex **then** *sentential*         (7)
    **elseif** position in intermediate phrase = ¬first **then** *sentential*         (8)

<u>Manually derived textual model (repeated from Figure 1):</u>

**if** preceding orthography = true **then** *discourse*         (9)
**elseif** preceding orthography = false **then** *sentential*         (10)

<u>Decision tree learned from speech-text using C4.5:</u>

**if** position in intonational phrase $\leq 1$ **then**
    **if** preceding orthography* = NA **then** *discourse*
    **elseif** preceding orthography* = true **then** *discourse*
    **elseif** preceding orthography* = false **then**
        **if** length of intermediate phrase $> 12$ **then** *discourse*
        **elseif** length of intermediate phrase $\leq 12$ **then**
            **if** length of intermediate phrase $\leq 1$ **then** *discourse*
            **elseif** length of intermediate phrase $> 1$ **then** *sentential*
**elseif** position in intonational phrase $> 1$ **then**
    **if** length of intermediate phrase $\leq 1$ **then** *discourse*
    **elseif** length of intermediate phrase $> 1$ **then** *sentential*

<u>Ruleset learned from speech-text using CGRENDEL:</u>

**if** (preceding orthography = false) $\wedge$ ($4 \leq$ position in intonational phrase $\leq 6$) $\wedge$ **then** *sentential*
**if** (preceding orthography = false) $\wedge$ (length of intermediate phrase $\geq 2$) **then** *sentential*
**if** (preceding orthography = false) $\wedge$ (length of intonational phrase $\geq 7$) $\wedge$ (preceding cue phrase = NA)
    $\wedge$ (accent = H*) **then** *sentential*
**if** (preceding orthography = comma) $\wedge$ (length of intermediate phrase $\geq 5$) $\wedge$ (length of intonational phrase $\leq 17$)
    $\wedge$ (part-of-speech = adverb) **then** *sentential*
**if** (preceding orthography = comma) $\wedge$ ($3 \leq$ length of intonational phrase $\leq 8$) $\wedge$ (accent = H*) **then** *sentential*
**if** (preceding orthography = comma) $\wedge$ ($3 \leq$ length of intermediate phrase $\leq 8$)
    $\wedge$ (length of intonational phrase $\geq 15$) **then** *sentential*
**if** (position in intonational phrase $\geq 2$) $\wedge$ (length of intermediate phrase $\geq 2$)
    $\wedge$ (preceding cue phrase = NA) **then** *sentential*
default is on *discourse*

Figure 7: C4.5 and CGRENDEL classification models learned from the prosodic/textual feature representation of the classifiable cue phrases in the multiple cue phrase corpus.

| Model | Classifiable Cue Phrases (N=878) | Classifiable Non-Conjuncts (N=495) |
|---|---|---|
| P-L+ | 21.8 ± 4.6 | 17.4 ± 2.7 |
| P-P+ | *16.7 ± 2.8* | 14.8 ± 5.0 |
| I-L+ | 20.3 ± 3.4 | 16.0 ± 3.3 |
| I-P+ | 25.1 ± 4.1 | 17.0 ± 3.6 |
| I-C+ | 27.0 ± 3.6 | 18.4 ± 3.4 |
| A+ | 19.8 ± 3.2 | 12.8 ± 3.1 |
| A*+ | 18.6 ± 3.8 | 15.4 ± 2.8 |
| prosody+ | *16.7 ± 2.9* | 15.8 ± 3.1 |
| hl93features+ | 24.0 ± 4.5 | 17.4 ± 4.3 |
| phrasing+ | *14.5 ± 3.3* | 12.6 ± 3.3 |
| length+ | *18.6 ± 2.0* | 16.2 ± 3.5 |
| position+ | *15.6 ± 3.3* | 13.0 ± 3.9 |
| intonational+ | *15.1 ± 2.2* | 16.6 ± 4.6 |
| intermediate+ | 18.5 ± 3.7 | 16.6 ± 4.0 |
| manual prosodic | 24.6 ± 3.0 | 14.7 ± 3.2 |

Table 8: 95%-confidence intervals for the error rates (%) of all CGRENDEL prosodic, *tokenized* classification models, testing data. (Training and testing were done from the multiple cue phrase corpus using cross-validation.)

Figure 2. These experiments investigate how performance changes when classification models are allowed to treat different cue phrases differently. As will be seen, learning from tokenized feature sets often further improves the performance of the learned classification models. In addition, the classification models now contain new linguistic information regarding particular tokens (e.g., "so").

### 4.3.1 Prosodic Models

Table 8 presents the error of the learned classification models on both test sets from the multiple cue phrase corpus, for each of the *tokenized* prosodic feature sets. Again, the error rates in italics indicate that the performance of the learned classification model meaningfully exceeds the performance of the "manual prosodic" model (which did not consider the feature *token*).

One way that the improvement obtained by adding the feature *token* can be seen is by comparing the performance of the learned and manually derived models. In Table 8, six CGRENDEL classification models have lower (italicized) error rates than the manual model. In Table 5, only five of these models are italicized. Thus, adding the feature *token* results in an additional learned model - *length+* - outperforming the manually derived model. Conversely, in Table 8, no learned models perform significantly worse than the manually derived manual. In contrast, in Table 5, several non-tokenized models perform worse than the manual model (I-C in the larger test set, and P-L, I-L, I-C, A, A*, and *length* in the non-conjunct test set).

The improvement obtained by adding the feature *token* can also be seen by comparing the performance of the tokenized (Table 8) and non-tokenized (Table 5) versions of each model to each other. For convenience, cases where tokenization yields improvement are highlighted in Table 9. The table shows that the error rate of the tokenized versions of the feature sets is significantly lower than the error of the non-tokenized versions, for P-L, I-C,

78

| Model | Classifiable Cue Phrases (N=878) | | Classifiable Non-Conjuncts (N=495) | |
|---|---|---|---|---|
| | Non-Tokenized | Tokenized (+) | Non-Tokenized | Tokenized (+) |
| P-L | $33.0 \pm 5.9$ | $21.8 \pm 4.6$ | $33.2 \pm 1.9$ | $17.4 \pm 2.7$ |
| I-L | - | - | $25.6 \pm 2.8$ | $16.0 \pm 3.3$ |
| I-C | $36.5 \pm 5.4$ | $27.0 \pm 3.6$ | $35.2 \pm 3.4$ | $18.4 \pm 3.4$ |
| A | $28.6 \pm 3.6$ | $19.8 \pm 3.2$ | $30.2 \pm 3.1$ | $12.8 \pm 3.1$ |
| A* | $28.3 \pm 4.3$ | $18.6 \pm 3.8$ | $28.4 \pm 1.7$ | $15.4 \pm 2.8$ |
| length | $26.1 \pm 3.8$ | $18.6 \pm 2.0$ | $27.4 \pm 3.4$ | $16.2 \pm 3.5$ |

Table 9: Cases where adding the feature *token* improves the performance of a prosodic model.

A, A*, and *length* in both test sets, and for I-L in only the non-conjunct test set. Note the overlap between the feature sets of Table 9 and those discussed in the previous paragraph.

Figure 8 shows several tokenized single feature prosodic classification models. The first CGRENDEL model in the figure shows the ruleset learned from P-L+, which reduces the $33.2\% \pm 1.9\%$ error rate of P-L (*length of intonational phrase*) to $17.4\% \pm 2.7\%$, when trained and tested using the classifiable non-conjuncts (Table 9). Note that the first rule uses only a prosodic feature (like the rules of Experiment Sets 1 and 2), and is in fact similar to line (1) of the manual model. (Recall that the length value 1 is equivalent to the composition value *alone*.) However, unlike the rules of the previous experiment sets, the next 5 rules use both the prosodic feature and the lexical feature *token*. Also unlike the rules of the previous experiment sets, the remaining rules classify cue phrases using only the feature *token*. Examination of the learned rulesets in Figures 8 and 9 shows that the same cue phrases often appear in this last type of rule. Some of these cue phrases, for example, "finally", "however", and "ok", are in fact always *discourse* usages in the multiple cue phrase corpus. For the other cue phrases, classifying cue phrases using only *token* corresponds to classifying cue phrases using their default class (the most frequent type of usage in the multiple cue phrase corpus). Recall the use of a non-tokenized default class model in Table 1.

The second example shows the ruleset learned from I-C+ (*composition of intermediate phrase+*). The first rule corresponds to line (1) of the manually derived model.[15] The next six rules classify particular cue phrases as *discourse*, independently of the value of I-C. Note that although in this model the cue phrase "say" is classified using only *token*, in the previous model a more sophisticated strategy for classifying "say" could be found.

The third example shows the CGRENDEL ruleset learned from A+ (*accent+*). The first rule corresponds to line (5) of the manually derived prosodic model. In contrast to line (4), however, CGRENDEL uses deaccenting to predict *discourse* for only the tokens "say" and "so." If the token is "finally", "however", "now" or "ok", *discourse* is assigned (for all accents). In all other deaccented cases, *sentential* is assigned (using the default). Similarly, in contrast to line (7), the complex accent L+H* predicts *discourse* for the cue phrases "further" and "indeed" (and also for "finally", "however", "now" and "ok"), and *sentential* otherwise.

---

15. As discussed in relation to Figure 2, the I-C values *only* and *only cue phrases* in the multiple cue phrase corpus replace the value *alone* in the "now" corpus.

Manually derived prosodic model (repeated from Figure 1):

| | |
|---|---|
| **if** composition of intermediate phrase = alone **then** *discourse* | (1) |
| **elseif** composition of intermediate phrase = ¬alone **then** | (2) |
|     **if** position in intermediate phrase = first **then** | (3) |
|         **if** accent = deaccented **then** *discourse* | (4) |
|         **elseif** accent = L* **then** *discourse* | (5) |
|         **elseif** accent = H* **then** *sentential* | (6) |
|         **elseif** accent = complex **then** *sentential* | (7) |
|     **elseif** position in intermediate phrase = ¬first **then** *sentential* | (8) |

Ruleset learned from P-L+ using CGRENDEL:

**if** length of intonational phrase ≤ 1 **then** *discourse*
**if** (7 ≤ length of intonational phrase ≤ 11) ∧ (token = although) **then** *discourse*
**if** (9 ≤ length of intonational phrase ≤ 16) ∧ (token = indeed) **then** *discourse*
**if** (length of intonational phrase ≤ 20) ∧ (token = say) **then** *discourse*
**if** (11 ≤ length of intonational phrase ≤ 13) ∧ (token = then) **then** *discourse*
**if** (length of intonational phrase = 5) ∧ (token = well) **then** *discourse*
**if** token = finally **then** *discourse*
**if** token = further **then** *discourse*
**if** token = however **then** *discourse*
**if** token = now **then** *discourse*
**if** token = ok **then** *discourse*
**if** token = otherwise **then** *discourse*
**if** token = so **then** *discourse*
default is on *sentential*

Ruleset learned from I-C+ using CGRENDEL:

**if** composition of intermediate phrase = only **then** *discourse*
**if** token = finally **then** *discourse*
**if** token = however **then** *discourse*
**if** token = now **then** *discourse*
**if** token = ok **then** *discourse*
**if** token = say **then** *discourse*
**if** token = so **then** *discourse*
default is on *sentential*

Ruleset learned from A+ using CGRENDEL:

**if** accent = L* **then** *discourse*
**if** (accent = deaccented) ∧ (token = say) **then** *discourse*
**if** (accent = deaccented) ∧ (token = so) **then** *discourse*
**if** (accent = L+H*) ∧ (token = further) **then** *discourse*
**if** (accent = L+H*) ∧ (token = indeed) **then** *discourse*
**if** token = finally **then** *discourse*
**if** token = however **then** *discourse*
**if** token = now **then** *discourse*
**if** token = ok **then** *discourse*
default is on *sentential*

Figure 8: Example CGRENDEL classification models learned from different *tokenized*, prosodic feature representations of the classifiable non-conjuncts in the multiple cue phrase corpus.

| Model | Classifiable Cue Phrases (N=878) | Classifiable Non-Conjuncts (N=495) |
|---|---|---|
| C-P+ | $(28.2 \pm 3.9)$ | $16.4 \pm 4.6$ |
| C-S+ | $(28.9 \pm 3.6)$ | $17.2 \pm 4.0$ |
| O-P+ | $17.5 \pm 4.4$ | $10.0 \pm 3.1$ |
| O-P*+ | $17.7 \pm 2.9$ | $12.2 \pm 2.9$ |
| O-S+ | $26.9 \pm 4.7$ | $18.4 \pm 3.9$ |
| O-S*+ | $(27.3 \pm 3.5)$ | $16.0 \pm 3.2$ |
| POS+ | $(27.4 \pm 3.6)$ | $17.2 \pm 3.9$ |
| text+ | $18.4 \pm 3.0$ | $12.0 \pm 2.6$ |
| adjacency+ | $(28.6 \pm 4.1)$ | $15.2 \pm 3.1$ |
| orthography+ | $17.6 \pm 3.0$ | $13.6 \pm 3.9$ |
| preceding+ | $17.0 \pm 4.1$ | $13.6 \pm 2.6$ |
| succeeding+ | $25.6 \pm 3.9$ | $18.0 \pm 4.5$ |
| manual textual | $19.9 \pm 2.8$ | $16.1 \pm 3.4$ |

Table 10: 95%-confidence intervals for the error rates (%) of all CGRENDEL textual, *tokenized* classification models, testing data. (Training and testing were done from the multiple cue phrase corpus using cross-validation.)

To summarize, new prosodic results of Experiment Set 3 are that features relating to length, composition, and accent, while not useful (in isolation) for predicting the classification of all cue phrases, are in fact quite useful for predicting the class of individual cue phrases or subsets of cue phrases. (Recall that the result of Experiment Sets 1 and 2 was that without *token*, only the prosodic feature *position in intonational phrase* was useful in isolation.)

### 4.3.2 Textual Models

Table 10 presents the error of the learned classification models on both test sets from the multiple cue phrase corpus, for each of the *tokenized* textual feature sets. As in Experiment Set 2 (Table 6), none of the CGRENDEL classification models have lower (italicized) error rates than the manual model. However, adding the feature *token* does improve the performance of many of the learned rulesets, in that the following models (unlike their non-tokenized counterparts) are no longer outperformed by the manual model: O-S+ and *succeeding+* in the larger test set, and C-P+, C-S+, O-S+, O-S*+, POS+, *adjacency+*, and *succeeding+* in the non-conjunct test set.

The improvement obtained by adding the feature *token* can also be seen by comparing the performance of the tokenized (Table 10) and non-tokenized (Table 6) versions of each model to each other, as shown in Table 11. The table shows that the error rates of the tokenized versions of the feature sets are significantly lower than the error of the non-tokenized versions, for C-P, C-S, POS, and *adjacency* in both test sets, and for O-P, O-S, O-S*, *text*, and *succeeding* in the non-conjunct test set. Note the overlap between the feature sets of Table 11 and those discussed in the previous paragraph.

Figure 9 shows several tokenized single textual feature classification models. The first CGRENDEL model shows the ruleset learned from C-P+ (*preceding cue phrase+*), which reduces the 40.2% ± 4.5% error rate of C-P to 16.4% ± 4.6% when trained and tested using the classifiable non-conjuncts (Table 11). This ruleset correlates preceding cue phrases with discourse usages of "indeed", and omitted transcriptions of "further", "now", and "so" with

Manually derived textual model (repeated from Figure 1):

**if** preceding orthography = true **then** *discourse*
**elseif** preceding orthography = false **then** *sentential*

Ruleset learned from C-P+ using CGRENDEL:

**if** (preceding cue phrase = true) ∧ (token = indeed) **then** *discourse*
**if** (preceding cue phrase = NA) ∧ (token = further) **then** *discourse*
**if** (preceding cue phrase = NA) ∧ (token = now) **then** *discourse*
**if** (preceding cue phrase = NA) ∧ (token = so) **then** *discourse*
**if** token = although **then** *discourse*
**if** token = finally **then** *discourse*
**if** token = however **then** *discourse*
**if** token = ok **then** *discourse*
**if** token = say **then** *discourse*
**if** token = similarly **then** *discourse*
default is on *sentential*

Ruleset learned from O-P+ using CGRENDEL:

**if** preceding orthography = false **then** *sentential*
**if** (preceding orthography = comma) ∧ (token = then) **then** *sentential*
default is on *discourse*

Ruleset learned from O-S+ using CGRENDEL:

**if** succeeding orthography = comma **then** *discourse*
**if** (succeeding orthography = false) ∧ (token = so) **then** *discourse*
**if** succeeding orthography = NA **then** *discourse*
**if** token = although **then** *discourse*
**if** token = finally **then** *discourse*
**if** token = now **then** *discourse*
**if** token = ok **then** *discourse*
**if** token = say **then** *discourse*
default is on *sentential*

Ruleset learned from POS+ using CGRENDEL:

**if** (part-of-speech = adverb) ∧ (token = finally) **then** *discourse*
**if** (part-of-speech = singular proper noun) ∧ (token = further) **then** *discourse*
**if** (part-of-speech = adverb) ∧ (token = however) **then** *discourse*
**if** (part-of-speech = adverb) ∧ (token = indeed) **then** *discourse*
**if** (part-of-speech = subordinating conjunction) ∧ (token = so) **then** *discourse*
**if** token = although **then** *discourse*
**if** token = now **then** *discourse*
**if** token = say **then** *discourse*
**if** token = ok **then** *discourse*
default is on *sentential*

Figure 9: Example CGRENDEL classification models learned from different *tokenized*, textual feature representations of the classifiable non-conjuncts in the multiple cue phrase corpus.

| Model | Classifiable Cue Phrases (N=878) | | Classifiable Non-Conjuncts (N=495) | |
|---|---|---|---|---|
| | Non-Tokenized | Tokenized (+) | Non-Tokenized | Tokenized (+) |
| C-P | 40.7 ± 6.2 | 28.2 ± 3.9 | 40.2 ± 4.5 | 16.4 ± 4.6 |
| C-S | 41.3 ± 5.9 | 28.9 ± 3.6 | 39.8 ± 4.2 | 17.2 ± 4.0 |
| O-P | - | - | 17.6 ± 3.3 | 10.0 ± 3.1 |
| O-S | - | - | 30.2 ± 1.8 | 18.4 ± 3.9 |
| O-S* | - | - | 32.6 ± 3.0 | 16.0 ± 3.2 |
| POS | 37.7 ± 4.1 | 27.4 ± 3.6 | 38.2 ± 4.6 | 17.2 ± 3.9 |
| text | - | - | 19.0 ± 3.6 | 12.0 ± 2.6 |
| adjacency | 39.7 ± 5.7 | 28.6 ± 4.1 | 40.2 ± 3.4 | 15.2 ± 3.1 |
| succeeding | - | - | 30.0 ± 2.7 | 18.0 ± 4.5 |

Table 11: Cases where adding the feature *token* improves the performance of a textual model.

discourse usages. The classifications for the rest of the cue phrases are predicted using only the feature token.

The second example shows the CGRENDEL ruleset learned from O-P+ (*preceding orthography+*). This ruleset correlates no preceding orthography with sentential usages of cue phrases (as in both the manually derived model and the learned models from Experiment Set 2). Unlike those models, however, the cue phrase "then" is also classified as *sentential*, even when it is preceded by orthography (namely, by a comma).

The third example shows the CGRENDEL ruleset learned from O-S+ (*succeeding orthography*). This ruleset correlates the presence of succeeding commas with discourse usages of cue phrases, except for the cue phrase "so", which is classified as a discourse usage without any succeeding orthography. The model also correlates cue phrases that were omitted from the transcript with discourse usages. The classifications for the rest of the cue phrases are predicted using only the feature *token*.

The last example shows the CGRENDEL ruleset learned from POS+ (*part-of-speech+*). This ruleset classifies certain cue phrases as discourse usages depending on both *part-of-speech* and *token*, as well as independently of *part-of-speech*.

Finally, Figure 10 shows the classification model learned from *text+*, the largest tokenized textual feature set. Note that three of the four features used in the tokenized, single textual feature models of Figure 9 are incorporated into this tokenized, multiple textual feature model.

To summarize, new textual results of Experiment Set 3 are that features based on adjacent cue phrases, succeeding orthography, and part-of-speech, while not useful (in isolation) for predicting the classification of all cue phrases, are in fact quite useful in conjunction with only the feature *token*. (Recall that the result of Experiment Set 2 was that without *token*, only the textual features *preceding orthography* and *preceding orthography\** were useful in isolation.)

### 4.3.3 Prosodic/Textual Models

Table 12 presents the error rates of the classification models learned by CGRENDEL when the data is represented using *speech-text+*, the complete set of prosodic and textual

Ruleset learned from text+ using CGRENDEL:

**if** preceding orthography = false **then** *sentential*
**if** (preceding orthography = comma) ∧ (token = although) **then** *sentential*
**if** (preceding orthography = comma) ∧ (token = no) **then** *sentential*
**if** (preceding orthography = comma) ∧ (token = then) **then** *sentential*
**if** (succeeding orthography = false) ∧ (preceding cue phrase = NA) ∧ (token = similarly) **then** *sentential*
**if** token = actually **then** *sentential*
**if** token = first **then** *sentential*
**if** token = since **then** *sentential*
**if** token = yes **then** *sentential*
default is on *discourse*

Figure 10: CGRENDEL classification model learned from a tokenized, *multiple* textual feature representation of the classifiable non-conjuncts in the multiple cue phrase corpus.

| Model | Classifiable Cue Phrases (N=878) | Classifiable Non-Conjuncts (N=495) |
|---|---|---|
| speech-text+ | *16.9 ± 3.4* | 16.6 ± 4.1 |
| manual prosodic | 24.6 ± 3.0 | 14.7 ± 3.2 |
| manual textual | 19.9 ± 2.8 | 16.1 ± 3.4 |

Table 12: 95%-confidence intervals for the error rates (%) of the CGRENDEL prosodic/textual, *tokenized* classification models, testing data. (Training and testing were done from the multiple cue phrase corpus using cross-validation.)

features. As in Experiment Set 2, the performance of *speech-text+* is not better than the performance of either the best learned (tokenized) prosodic or textual models (Tables 8 and 10, respectively).

Comparison of Tables 7 and 12 also shows that for the feature set *speech-text*, tokenization does not improve performance. This is in contrast to the prosodic and textual feature sets, where tokenization improves the performance of many learned models (namely those shown in Tables 9 and 11).

### 4.4 Experiment Set 4: Adding the Classification *ambiguous*

In practice, a cue phrase classification model will have to classify *all* the cue phrases in a recording or text, not just those that are "classifiable." The experiments in the fourth set replicate the experiments in Experiment Sets 1, 2, and 3, with the exception that all 953 cue phrases in the multiple cue phrase corpus are now used. This means that cue phrases are now classified as *discourse*, *sentential*, as well as *unknown* (defined in Table 2). Experiment Set 4 investigates whether machine learning can explicitly recognize the new class *unknown*.

Recall that the studies of Hirschberg and Litman did not attempt to predict the class *unknown*, as it did not occur in their "now" training corpus. Thus in Experiment Set 1, the class *unknown* similarly can not be learned from the training data. However, the *unknown* examples can be added to the testing data of Experiment Set 1. Obviously performance will degrade, as the models must incorrectly classify each *unknown* example as either *discourse* or *sentential*. For example, when tested on the full corpus of 953 example cue phrases, the 95% confidence intervals for the error rates of P-P and *intonational* are 24.8% ± 2.8%; recall that when tested on the subset of the corpus corresponding to the 878 classifiable cue phrases, the error was 18.3% ± 2.6% (Table 4).

Unfortunately, the results of rerunning Experiment Sets 2 and 3 do not show promising results for classifying cue phrases as *unknown*. Despite the presence of 75 examples of *unknown*, most of the learned models still classify *unknown* cue phrases as only *discourse* or *sentential*. For example, when CGRENDEL is used for learning, only 2 of the possible 27 non-tokenized models[16] (*phrasing* and *speech-text*) contain rules that predict the class *unknown*. Furthermore, each of these models only contains one rule for *unknown*, and each of these rules only applies to 2 of the possible 953 examples! Similarly, only four of the possible 27 tokenized models (*length+*, *phrasing+*, *prosody+*, and *speech-text+*) contain at least one rule for the class *unknown*. When compared to training and testing using only the classifiable cue phrases in the corpus, the error rate on the full corpus is typically (but not always) significantly higher. The best performing model in Experiment Set 4 is *speech-text+*, with a 22.4% ± 4.1% error rate (95% confidence interval).

In sum, Experiment Set 4 addressed a problem that was previously unexplored in the literature - the ability to develop classification models that predict not only *discourse* and *sentential* usages of cue phrases, but also usages which human judges find difficult to classify. Unfortunately, the results of the experiments suggest that learning how to classify cue phrases as *unknown* is a difficult problem. Perhaps with more training data (recall that there are only 75 examples of *unknown*) or with additional features better results could be obtained.

## 4.5 Discussion

The experimental results suggest that machine learning is a useful tool for both automating the generation of classification models and improving upon manually derived results. In Experiment Sets 1 and 2 the performance of many of the learned classification models is comparable to the performance of the manually derived models. In addition, when tested on the classifiable cue phrases, several learned prosodic classification models (as well as the learned prosodic/textual model) outperform Hirschberg and Litman's manually derived prosodic model. Experiment Set 3 shows that learning from tokenized feature sets even further improves performance, especially in the non-conjunct test set. More tokenized than non-tokenized learned models perform at least as well as the manually derived models. Many tokenized learned models also outperform their non-tokenized counterparts.

While the textual classification models do not outperform the better prosodic classification models, they have the advantage that the textual feature values are obtained directly from the transcript, while determining the values of prosodic features requires manual analysis. (See, however, Section 5 for a discussion of the feasibility of automating the prosodic analysis. In addition, a transcript may not always be available.) On the other hand, almost all the high performing textual models are dependent on orthography. While manual tran-scriptions of prosodic features have been shown to be reliable across coders (Pitrelli et al., 1994), there are no corresponding results for the reliability of orthography.

---

16. Recall that Experiment Sets 2 and 3 constructed 14 prosodic models, 12 textual models, and 1 prosodic/textual model.

Examination of the best performing learned models shows that they are often comparable in content to the relevant portions of the manually derived models. Examination of the models also provides new contributions to the cue phrase literature. For example, Experiment Sets 1 and 2 demonstrate the utility of classifying cue phrases based on only a single prosodic feature - phrasal position.[17] Experiment Set 2 also demonstrates the utility of the prosodic feature *length* and the textual feature *preceding cue phrase* for classifying cue phrases - in conjunction with other prosodic and textual features. Finally, the results of Experiment Set 3 demonstrate that even though many features are not useful by themselves for classifying all cue phrases, they may nonetheless be very informative in their *tokenized* form. This is true for the prosodic features based on phrasal length, phrasal composition, and accent, and for the textual features based on adjacent cue phrases, succeeding position, and part-of-speech.[18]

## 5. Utility

The results of the machine learning experiments are quite promising, in that when compared to manually derived classification models already in the literature, the learned classification models often perform with comparable if not higher accuracy. Thus, machine learning appears to be an effective technique for automating the generation of classification models. However, given that the experiments reported here still rely on manually created training data, a discussion of the practical utility of the results is in order.

Even given manually created training data, the results established by Hirschberg and Litman (1993) - obtained using even less automation than the experiments of this paper - are already having practical import. In particular, the manually derived cue phrase classification models are used to improve the naturalness of the synthetic speech in a text-to-speech system (Hirschberg, 1990). Using the text-based model, the text-to-speech system classifies each cue phrase in a text to be synthesized as either a discourse or sentential usage. Using the prosodic model, the system then conveys this usage by synthesizing the cue phrase with the appropriate type of intonation. The speech synthesis could be further improved (and the output made more varied) by using any one of the higher performing learned prosodic models presented in this paper.

The results of this paper could also be directly applied in the area of text generation. For example, Moser and Moore (1995) are concerned with the implementation of cue selection and placement strategies in natural language generation systems. Such systems could be enhanced by using the text-based models of cue phrase classification (particularly the tokenized models) to additionally specify preceding and succeeding orthography, part-of-speech, and adjacent cue phrases that are appropriate for discourse usages.

---

17. The empirical studies performed by Holte (1993) show that for many other datasets, the accuracy of single feature rules and decision trees is often competitive with the accuracy of more complex learned models.

18. In contrast, the prosodic features phrasal composition and accent were previously known to be useful *in conjunction with* each other and with phrasal position (Hirschberg & Litman, 1993), while part-of-speech was known to be useful only in conjunction with orthography (Hirschberg & Litman, 1993). Length, adjacent cue phrases, and succeeding position were not used in either of the manually derived models (Hirschberg & Litman, 1993) (although length and adjacent cue phrases were shown to be useful - again only in conjunction with other prosodic and textual features - in Experiment Set 2).

Finally, if the results of this paper could be fully automated, they could also be used in natural language *understanding* systems, by enhancing their ability to recognize discourse structure. The results obtained by Litman and Passonneau (1995) and Passonneau and Litman (in press) suggest that algorithms that use cue phrases (in conjunction with other features) to predict discourse structure outperform algorithms that do not take cue phrases into account. In particular, Litman and Passonneau develop several algorithms that explore how features of cue phrases, prosody and referential noun phrases can be best combined to predict discourse structure. Quantitative evaluations of their results show that the best performing algorithms all incorporate the use of discourse usages of cue phrases (where cue phrases are classified as discourse using only phrasal position). As discussed in Section 1, discourse structure is useful for performing tasks such as anaphora resolution and plan recognition. Recent work has also shown that if discourse structure can be recognized, it can be used to improve retrieval of text (Hearst, 1994) and speech (Stifleman, 1995).

Although the prosodic features were manually labeled by Hirschberg and Litman, there are recent results suggesting that at least some aspects of prosody can be automatically labeled directly from speech. For example, Wightman and Ostendorf (1994) develop an algorithm that is able to automatically recognize prosodic phrasing with 85-86% accuracy (measured by comparing automatically derived labels with hand-marked labels); this accuracy is only slightly less than human-human accuracy. Recall that the experimental results of this paper show that models learned from the single feature *position in intonational phrase* - which could be automatically computed given such an automatic prosodic phrasing algorithm - perform at least as well as any other learned prosodic model. Similarly, accenting versus deaccenting can be automatically labeled with 88% accuracy (Wightman & Ostendorf, 1994), while a more sophisticated labeling scheme that distinguishes between four types of accent classes (and is somewhat similar to the prosodic feature *accent\** used in this paper) can be labeled with 85% accuracy (Ostendorf & Ross, in press). Recall from Experiment Set 3 that the tokenized models learned using *accent\** also classify cue phrases with good results.

Although the textual features were automatically extracted from a transcript, the transcript itself was manually created. Many natural language understanding systems do not deal with speech at all, and thus begin with such textual representations. In spoken language systems the transcription process is typically automated using a speech recognition system (although this introduces further sources of error).

## 6. Related Work

This paper has both compared the results obtained using machine learning to previously existing manually-obtained results, and has also used machine learning as a tool for developing theories given new linguistic data (as in the models resulting from Experiment Set 3, where the new feature *token* was considered). Siegel (1994) similarly uses machine learning (in particular, a genetic learning algorithm) to classify cue phrases from a previously unstudied set of textual features: a feature corresponding to *token*, as well as textual features containing the lexical or orthographic item immediately to the left of and in the 4 positions to the right of the example. Siegel's input consists of one judge's non-ambiguous examples taken from the data used by Hirschberg and Litman (1993) as well as additional examples; his output is in the form of decision trees. Siegel reports a 21% estimated error rate, with half of the corpus used for training and half for testing. Siegel and McKeown (1994) also propose a method for developing linguistically viable rulesets, based on the partitioning of the training data produced during induction.

Machine learning has also been used in several other areas of discourse analysis. For example, learning has been used to develop rules for structuring discourse into multi-utterance segments. Grosz and Hirschberg (1992) use the classification and regression tree system CART (Brieman et al., 1984) to construct decision trees for classifying aspects of discourse structure from intonational feature values. Litman and Passonneau (1995) and Passonneau and Litman (in press) use the system C4.5 to construct decision trees for classifying utterances as discourse segment boundaries, using features relating to prosody, referential noun phrases, and cue phrases. In addition, C4.5 has been used to develop anaphora resolution algorithms, by training on corpora tagged with appropriate discourse information (Aone & Bennett, 1995). Similarly, McCarthy and Lehnert (1995) use C4.5 to learn decision trees to classify pairs of phrases as coreferent or not. Soderland and Lehnert (1994) use the machine learning program ID3 (a predecessor of C4.5) to support corpus-driven knowledge acquisition in information extraction. Machine learning often results in algorithms that outperform manually derived alternatives (Litman & Passonneau, 1995; Passonneau & Litman, in press; Aone & Bennett, 1995; McCarthy & Lehnert, 1995), although statistical inference is not always used to evaluate the significance of the performance differences.

Finally, machine learning has also been used with great success in many other areas of natural language processing. As discussed above, the work of most researchers in discourse analysis has concentrated on the direct application of existing symbolic learning approaches (e.g., C4.5), and on the comparison of learning and manual methods. While researchers in other areas of natural language processing have also addressed these issues, they have in addition applied a much wider variety of learning approaches, and have been concerned with the development of learning methods particularly designed for language processing. A recent survey of learning for natural language (Wermter, Riloff, & Scheler, 1996) illustrates both the type of learning approaches that have been used and modified (in particular, symbolic, connectionist, statistical, and hybrid approaches), as well as the scope of the problems that have proved amenable to the use of learning techniques (e.g., grammatical inference, syntactic disambiguation, and word sense disambiguation).

## 7. Conclusion

This paper has demonstrated the utility of machine learning techniques for cue phrase classification. Machine learning supports the automatic generation of linguistically viable classification models. When compared to manually derived models already in the literature, many of the learned models contain new linguistic insights and perform with at least as high (if not higher) accuracy. In addition, the ability to automatically construct classification models makes it easier to comparatively analyze the utility of alternative feature representations of the data. Finally, the ease of retraining makes the learning approach more scalable and extensible than manual methods.

A first set of experiments were presented that used the machine learning programs CGRENDEL (Cohen, 1992, 1993) and C4.5 (Quinlan, 1993) to induce classification models from the preclassified cue phrases and their features that were used as training data by Hirschberg and Litman (1993). These results were then evaluated with the same testing data and methodology used by Hirschberg and Litman (1993). A second group of experiments used the method of cross-validation to both train and test from the testing data used by Hirschberg and Litman (1993). A third set of experiments induced classification models using the new feature *token*. A fourth set of experiments induced classification models using the new classification *unknown*.

The experimental results indicate that several learned classification models (including extremely simple one feature models) have significantly lower error rates than the models developed by Hirschberg and Litman (1993). One possible explanation is that the hand-built classification models were derived using very small training sets; as new data became available, this data was used for testing but not for updating the original models. In contrast, machine learning in conjunction with cross-validation (Experiment Set 2) supported the building of classification models using a much larger amount of the data for training. Even when the learned models were derived using the same small training set (Experiment Set 1), the results showed that the learning approach helped guard against overfitting on the training data.

While the prosodic classification model developed by Hirschberg and Litman demonstrated the utility of combining phrasal position with phrasal composition and accent, the best performing prosodic models of Experiment Sets 1 and 2 demonstrated that phrasal position was in fact even more useful for predicting cue phrases when used by itself. The other high performing classification models of Experiment Set 2 also demonstrated the utility of classifying cue phrases based on the prosodic feature *length* and the textual feature *preceding cue phrase*, in combination with other features.

Just as the machine learning approach made it easy to retrain when new training examples became available (Experiment Set 2), machine learning also made it easy to retrain when new features become available. In particular, when the value of the feature *token* was added to all the representations in Experiment Set 2, it was trivial to relearn all of the models (Experiment Set 3). Allowing the learning programs to treat cue phrases individually further improved the accuracy of the learned classification models, and added to the body of linguistic knowledge regarding cue phrases. Experiment Set 3 demonstrated that while not useful by themselves for classifying all cue phrases, the prosodic features based on phrasal length, phrasal composition, and accent, and textual features based on adjacent cue phrases, succeeding position, and part-of-speech, were in fact useful when used only in conjunction with the feature *token*.

A final advantage of the machine learning approach is that the ease of inducing classification models from many different sets of features supports an exploration of the comparative utility of different knowledge sources. This is especially useful for understanding the trade-offs between the accuracy of a model and the set of features that are considered. For example, it might be worth the effort to code a feature that is not automatically obtainable or that is expensive to automatically obtain if adding the feature results in a significant improvement in performance.

In sum, the results of this paper suggest that machine learning is a useful tool for cue phrase classification, when the amount of data precludes effective human analysis, when the flexibility afforded by easy retraining is needed (e.g., due to additional training examples, new features, new classifications), and/or when an analysis goal is to gain a better understanding of the different aspects of the data.

Several areas for future work remain. First, there is still room for performance improvement. The error rates of the best performing learned models, even though they outperform the manually derived models, perform with error rates in the teens. Note that only the features that were coded or discussed by Hirschberg and Litman (1993) were considered in this paper. It may be possible to further lower the error rates by considering new types of prosodic and textual features (e.g., other contextual textual features (Siegel, 1994), or features that have been proposed in connection with the more general topic of discourse structure), and/or by using different kinds of learning methods. Second, Experiment Set 4 (and the previous literature) show that as yet, there are no models for predicting when a cue phrase usage should be classified as *unknown*, rather than as *discourse* or *sentential*. Again, it may be possible to improve the performance of the existing learned models by considering new features and/or learning methods, or perhaps performance could be improved by providing more training data. Finally, it is currently an open question whether the textual models developed here, which were based on transcripts of speech, are applicable to written texts. Textual models thus need to be developed using written texts as training data. Machine learning should continue to be a useful tool for helping to address these issues.

## Appendix A. C4.5 Results for Experiment Sets 2 and 3

Tables 13, 14 and 15 present the C4.5 error rates for Experiment Sets 2 and 3. The C4.5 results for Experiment Set 2 are shown in the "Non-Tokenized" columns. A comparison of Tables 13 and 5 shows that except for A in the larger test set, the C4.5 prosodic error rates fall within the CGRENDEL confidence intervals. A similar comparison of Tables 14 and 6 shows that except for O-P in the larger test set, the C4.5 textual error rates fall within the CGRENDEL confidence intervals. Finally, a comparison of Tables 15 and 7 shows that the C4.5 error rate of *speech-text* falls within the CGRENDEL confidence interval. The fact that comparable CGRENDEL and C4.5 results are generally obtained suggests that the ability to automate as well as to improve upon manual performance is not due to the specifics of either learning program.

The C4.5 results for Experiment Set 3 are shown in the "Tokenized" columns of Tables 13, 14 and 15. Comparison with Tables 8, 10 and 12 shows that the error rates of C4.5 and CGRENDEL are not as similar as in Experiment Set 2. However, the error rates reported in the tables use the default C4.5 and CGRENDEL options when running the learning programs. Comparable performance between the two learning programs can in fact generally be achieved by overriding one of the default C4.5 options. As detailed by Quinlan (1993), the default C4.5 approach – which creates a separate subtree for each possible feature value – might not be appropriate when there are many values for a feature. This situation characterizes the feature *token*. When the C4.5 default option is changed to allow feature values to be grouped into one branch of the decision tree, the problematic C4.5 error rates do indeed improve. For example, the A+ error rate for the classifiable non-conjuncts changes from 29.2% (Table 13) to 11%, which is within the 12.8% ± 3.1% CGRENDEL confidence interval (Table 8).

| Model | Classifiable Cue Phrases (N=878) | | Classifiable Non-Conjuncts (N=495) | |
|---|---|---|---|---|
| | Non-Tokenized | Tokenized (+) | Non-Tokenized | Tokenized (+) |
| P-L | 32.5 | 31.7 | 32.2 | 31.4 |
| P-P | 16.2 | 18.4 | 18.8 | 19.0 |
| I-L | 25.6 | 26.8 | 25.6 | 25.6 |
| I-P | 25.9 | 26.3 | 19.4 | 18.8 |
| I-C | 36.5 | 36.6 | 35.8 | 32.8 |
| A | 40.7 | 40.7 | 29.6 | 29.2 |
| A* | 28.3 | 26.7 | 28.8 | 31.2 |
| prosody | 16.0 | 15.2 | 19.4 | 16.0 |
| hl93features | 30.2 | 29.0 | 18.8 | 18.8 |
| phrasing | 15.9 | 15.2 | 18.0 | 17.4 |
| length | 24.8 | 24.4 | 26.2 | 24.2 |
| position | 18.1 | 18.0 | 19.6 | 17.6 |
| intonational | 16.8 | 16.6 | 18.8 | 19.8 |
| intermediate | 21.2 | 22.3 | 21.6 | 18.4 |

Table 13: Error rates (%) of the C4.5 prosodic classification models, testing data. (Training and testing were done from the multiple cue phrase corpus using cross-validation.)

| Model | Classifiable Cue Phrases (N=878) | | Classifiable Non-Conjuncts (N=495) | |
|---|---|---|---|---|
| | Non-Tokenized | Tokenized (+) | Non-Tokenized | Tokenized (+) |
| C-P | 40.7 | 39.3 | 39.2 | 33.6 |
| C-S | 40.7 | 39.9 | 39.2 | 39.2 |
| O-P | 40.7 | 35.7 | 18.6 | 14.6 |
| O-P* | 18.4 | 20.3 | 17.2 | 15.0 |
| O-S | 35.0 | 31.6 | 31.8 | 31.8 |
| O-S* | 34.4 | 32.5 | 31.0 | 32.4 |
| POS | 40.7 | 34.7 | 41.8 | 31.8 |
| text | 19.0 | 20.6 | 20.0 | 15.0 |
| adjacency | 40.9 | 39.4 | 40.6 | 43.6 |
| orthography | 18.9 | 19.3 | 17.8 | 18.0 |
| preceding | 18.7 | 19.3 | 19.2 | 16.0 |
| succeeding | 34.1 | 32.9 | 30.0 | 31.8 |

Table 14: Error rates (%) of the C4.5 textual classification models, testing data. (Training and testing were done from the multiple cue phrase corpus using cross-validation.)

## Acknowledgements

I would like to thank William Cohen and Jason Catlett for their helpful comments regarding the use of CGRENDEL and C4.5, and Sandra Carberry, Rebecca Passonneau, and the three anonymous JAIR reviewers for their helpful comments on this paper. I would also like to thank William Cohen, Ido Dagan, Julia Hirschberg, and Eric Siegel for comments on a preliminary version of this paper (Litman, 1994).

| Model | Classifiable Cue Phrases (N=878) | | Classifiable Non-Conjuncts (N=495) | |
|---|---|---|---|---|
| | Non-Tokenized | Tokenized (+) | Non-Tokenized | Tokenized (+) |
| speech-text | 15.3 | 13.6 | 16.8 | 17.6 |

Table 15: Error rates (%) of the C4.5 prosodic/textual classification model, testing data. (Training and testing were done from the multiple cue phrase corpus using cross-validation.)